\documentclass[lettersize,journal]{IEEEtran}
\usepackage{amsmath,amsfonts}
\usepackage{algorithmicx}
\usepackage{array}
\usepackage{textcomp}
\usepackage{stfloats}
\usepackage{url}
\usepackage{verbatim}
\usepackage{graphicx}
\usepackage{hyperref}
\def\BibTeX{{\rm B\kern-.05em{\sc i\kern-.025em b}\kern-.08em
    T\kern-.1667em\lower.7ex\hbox{E}\kern-.125emX}}
\usepackage{balance}
\usepackage{cite}
\usepackage{enumitem}
\usepackage{algorithm}
\usepackage{algpseudocode}
\usepackage{booktabs}
\usepackage{multirow}
\usepackage{subfigure}
\hypersetup{
	colorlinks=true,
	linkcolor=blue,   
	citecolor=blue,     
	urlcolor=magenta,  
}
\usepackage{xcolor}
\begin{document}
\title{SCDM-HTD: Spectral-Wise Conditional Diffusion Model for Hyperspectral Target Detection}
	\author{Dunbin~Shen,
	Wenfeng~Kong,
	Jianjun~Liu,~\IEEEmembership{Member,~IEEE},
	Xiaorui~Ma,~\IEEEmembership{Member,~IEEE}
	Hongyu~Wang,~\IEEEmembership{Member,~IEEE},
	\thanks{This work was supported by the National Natural Science Foundation of China under Grant 62271102. \textit{(Corresponding author: Xiaorui~Ma.)}
	}
	\thanks{Dunbin Shen, Wenfeng Kong, Xiaorui Ma, and Hongyu Wang are with the School of Information and Communication Engineering, Dalian University of Technology, Dalian 116024, China (e-mail: sdb\_2012@163.com; kksamakwf@163.com; maxr@dlut.edu.cn; whyu@dlut.edu.cn).}
	\thanks{Jianjun Liu is with the School of Artificial Intelligence and Computer Science, Jiangnan University, Wuxi 214122, China (e-mail: liuofficial@163.com).}
}

%\markboth{Journal of \LaTeX\ Class Files,~Vol.~xx, No.~xx, December~2024}%
%{How to Use the IEEEtran \LaTeX \ Templates}

\maketitle
%However, this task faces challenges such as dependence on background priors and sensitivity to spectral variation, both of which may reduce detection reliability and robustness.
\begin{abstract}
	Hyperspectral target detection (HTD) aims to  locate and recognize objects at a fine-grained scale, playing a pivotal role in remote sensing. Despite significant progress, current HTD methods primarily focus on feature comparison while neglecting distribution learning for background-target separation, which can limit detection robustness in practical applications. This paper addresses this issue by introducing a spectral-wise conditional diffusion model that reformulates HTD as a task of generating a detection map. Specifically, we treat background and target confidence scores as training samples and use the corresponding pixel spectra as guiding conditions to capture the distributional discrepancy between background and target through diffusion and denoising. To enhance practical applicability, we propose a  lightweight conditional denoising network based on multilayer perceptron to map noisy confidence scores to their ideal counterparts and adopt a deterministic reverse non-Markov chain to skip sampling steps, thereby accelerating the generation process. Furthermore, we introduce a composite loss to improve background-target separation. This loss integrates constrained energy minimization loss to enhance target saliency and suppress the background by maximizing the target response while minimizing average output energy, alongside binary cross-entropy loss to classify background and target using pixel-wise labels. Experimental results on four benchmark datasets demonstrate the effectiveness of the proposed method, with improvements in both quantitative metrics and qualitative visual effects. The code is available at \url{https://github.com/shendb2022/SCDM-HTD}.
\end{abstract}
\begin{IEEEkeywords}
	Hyperspectral target detection, distribution learning,  diffusion model, conditional denoising
\end{IEEEkeywords}

\section{Introduction}
The advancement of remote sensing technology has propelled the development of precision Earth observation. Hyperspectral imaging is a widely used technique that captures three-dimensional data integrating both spectral and spatial information. The resulting hyperspectral image (HSI) comprises two spatial dimensions, which depict the geometric structure of a scene, and one spectral dimension, which characterizes the electromagnetic radiation of materials \cite{sun2024applications, Lei2021deep, wang2021review, Xu2021hyper, fang2024wide}. This spectral information encodes distinct physical and chemical properties, enabling precise identification of ground objects \cite{plaza2009recent, Dong2023joint, Li2024local, Zhang2024multi}. Hyperspectral target detection (HTD) is a key task in HSI processing, aimed at locating and identifying targets of interest within a complex background at the pixel or subpixel level \cite{chen2023target}. Compared to traditional object detection, HTD enables fine-grained target identification, even in scenarios with low spatial resolution and image corruption. As a result, HTD has been widely applied in various fields, including camouflage detection \cite{hupel2022adopting}, forest fire monitoring \cite{dennison2009daytime}, disaster response \cite{eismann2009automated}, and urban and rural planning \cite{abbate2003evaluation}.

Based on the availability of target priors, HTD can be categorized into target detection and anomaly detection. Target detection identifies pixels that match a given target prior, enabling high detection accuracy \cite{sneha2022hyperspectral}. In contrast, anomaly detection locates pixels that deviate from the surrounding background, operating in a fully unsupervised manner, which enhances its practicality \cite{su2021hyperspectral}. This paper focuses on target detection, where a single target spectrum serves as prior knowledge. Over the past decades, various detection methods have been developed. Statistical-based methods \cite{ kraut2005adaptive, chang2005orthogonal, farrand1997mapping} typically assume that the background follows a Gaussian distribution or adheres to an energy minimization principle, detecting targets by computing statistical measures. While simple and efficient, these methods may struggle in complex real-world scenarios due to mismatches with prior assumptions. Representation-based methods \cite{li2015combined, zhu2023learning, zhao2022hyper} formulate HTD as an optimization problem, estimating representation components via matrix or tensor decomposition with appropriate regularizations. These methods offer physical and mathematical interpretability but may suffer from handcrafted regularization inaccuracies and limited nonlinear representation capabilities. Deep learning-based methods \cite{shi2020sensor, rao2022siamese, wang2023self} leverage neural networks to extract discriminative features. To mitigate the challenge of limited labeled samples, strategies such as transfer learning \cite{shi2020sensor}, data augmentation \cite{rao2022siamese}, and unsupervised learning \cite{wang2023self} have been explored. However, these data-driven methods often rely on deep network architectures, which may introduce redundancy and lack physical interpretability. Furthermore, the absence of explicit data distribution learning can compromise detection robustness in real-world applications.

To capture data distribution for background-target discrimination, several distribution learning-based methods have been explored. For example, Xie \emph{et al.} \cite{xie2020background} proposed a background reconstruction model that estimates the background distribution using an adversarial autoencoder (AAE) with a target suppression constraint. Gao \emph{et al.} \cite{gao2023hyper} leveraged a generative adversarial network (GAN) as an auxiliary tool to augment background and target samples, which were then used to train the detector. To address the challenge of limited prior knowledge and better exploit background information, Qin \emph{et al.} \cite{qin2023weakly} introduced a Wasserstein GAN-based approach that incorporates weakly supervised learning and channel attention to estimate the background and generate detection proposals. Despite these advancements, distribution learning-based methods largely rely on autoencoders (AEs) or GANs, which suffer from weak generative capabilities and training instability issues such as mode collapse and gradient vanishing.

Recently, diffusion models have gained significant attention for their ability to generate high-quality and realistic data. The denoising diffusion probabilistic model (DDPM) \cite{NEURIPS2020_4c5bcfec}, a cornerstone in this family, extends the variational autoencoder by introducing multiple latent variables to model progressive probability transitions via a Markov chain. It corrupts data by incrementally adding Gaussian noise and trains a neural network to reverse this process. New samples are generated by starting from Gaussian noise and progressively denoised through the reverse Markov chain. Leveraging this powerful generative capability, DDPM has been widely applied across various tasks. However, its application in HTD remains limited. Existing studies primarily use diffusion models as auxiliary tools for background-target discrimination. Some estimate background distributions for suppression \cite{ma2023bsdm, Chen2025dwsdiff}, and some generate diverse HSIs to mitigate spectral variation \cite{Li2024ajoint}. These indirect approaches do not treat HTD as a generative task, and their non-end-to-end frameworks may constrain detection accuracy and efficiency. Therefore, an efficient diffusion model that directly captures the distributional discrepancy between background and target is needed for improved detection.

To address the aforementioned issues, this paper proposes a spectral-wise conditional diffusion model for HTD (SCDM-HTD). The model estimates the distribution of confidence scores using corresponding pixel spectra as guiding conditions and directly generates the detection map. Detection is achieved by distinguishing between the reconstructed background and target confidence scores. Specifically, to obtain sufficient background and target samples, we treat the original pixels as background and generate target samples by mixing them with the target prior, simulating spectral mixture and variation. The corresponding confidence scores are computed using cosine similarity between the pixels and the target prior. We then introduce the spectral conditional diffusion model to estimate the confidence score distribution. In the forward process, confidence score pairs from the same position are progressively diffused into noisy sequences by adding Gaussian noise. To efficiently recover the confidence score pairs, we design a lightweight multilayer perceptron (MLP)-based conditional denoising network (CDNet), which embeds pixel spectra as guiding conditions. A tailored background-target separation loss supervises the denoising process, integrating constrained energy minimization (CEM) loss to enhance target saliency while suppressing the background, and binary cross-entropy (BCE) loss to enforce binary classification of background and target. In the reverse process, the detection map is progressively refined from Gaussian noise, with the original HSI serving as the guiding condition. To accelerate the generation process, we adopt a deterministic reverse non-Markov chain, inspired by the denoising diffusion implicit model (DDIM) \cite{song2020denoising}, enabling step skipping during sampling. This acceleration significantly reduces the required number of steps while maintaining promising detection performance, facilitating practical applications. Experimental results on four benchmark datasets validate the effectiveness of the proposed SCDM-HTD. In summary, the main contributions of this paper are as follows:
\begin{enumerate}
	\item We propose a spectral-wise conditional model to transform HTD into a task of generating a detection map. It employs the background and target confidence scores as negative and positive samples, respectively, and the corresponding pixel spectra as guiding conditions, to capture the distributional discrepancy between background and target through diffusion and denoising. Thus, the desired confidence map can be refined via iterative sampling to accomplish detection.
    \item We propose a lightweight CDNet to model the mapping between noisy confidence scores and the ideal ones, employing MLPs to embed spectral conditions and extract distributional features. Additionally, we adopt a deterministic reverse non-Markov chain to skip sampling steps and accelerate the generation process. Compared to DDPM, the parameter capacity and computational requirements are significantly reduced, facilitating practical application.
    \item We propose a composite background-target separation loss to distinguish between background and target, and enhance detection robustness. It integrates CEM loss to highlight the target and suppress the background by maximizing the target response while minimizing the average output energy, and BCE loss to classify background and target into two classes using pixel-wise labels.
\end{enumerate}

The remainder of the paper is organized as follows: Section \ref{related_work} surveys existing HTD methods and related diffusion models. Section \ref{method} presents the proposed method in detail. Section \ref{experiments} provides experimental results and discussions. Finally, Section \ref{conclusion} concludes the paper.
\section{Related Work} 
\label{related_work}
\subsection{HTD Methods Review}
Existing HTD methods can be broadly classified into three categories: statistical-based methods, representation-based methods, and deep learning-based methods.
\subsubsection{Statistical-Based Methods}
These methods typically assume that the background or target follows specific distributions or principles to compute relevant statistics for detection. Classical methods such as the generalized likelihood ratio test (GLRT) \cite{kelly1986adaptive}, the adaptive coherence estimator (ACE) \cite{kraut2005adaptive}, and the adaptive matched filter (AMF) \cite{fuhrmann1992cfar} operate under the assumption that the background follows a multivariate Gaussian distribution, detecting targets by evaluating the likelihood ratio. Under the same Gaussian assumption, the orthogonal subspace projection (OSP) method \cite{chang2005orthogonal} detects targets by first projecting each pixel onto the orthogonal subspace of the background and then onto the target space to enhance the signal-to-noise ratio. Beyond distributional assumptions, some methods leverage energy minimization principles. For example, the CEM method \cite{farrand1997mapping} employs a finite impulse response filter that suppresses the background while enhancing the target by minimizing the average output energy and maximizing the target response. To improve generalization, various CEM-based extensions have been developed. Kernel CEM (KCEM) \cite{fuhrmann1992cfar} maps data into a high-dimensional kernel space to enhance linear separability. Inspired by neural networks, hierarchical CEM (hCEM) \cite{zou2015hierarchical} and ensemble-based cascaded CEM (E-CEM) \cite{zhao2019ensemble} employ multilayer architectures with CEM as the fundamental module, incorporating nonlinear activation functions to improve representation. To address CEM’s sensitivity to the choice of data origin (centralization points), the clever eye (CE) method \cite{geng2016clever} treats the data origin as an optimization variable and derives a global closed-form solution. While statistical-based methods are efficient and effective for many HTD tasks, their performance can be limited in real-world datasets due to mismatches between prior assumptions and the complex, non-Gaussian nature of practical distributions.
\subsubsection{Representation-Based Methods}
These methods formulate HTD as an optimization problem, aiming to estimate representation components through matrix or tensor decomposition, combined with various regularization techniques. Matrix decomposition-based methods typically rely on a linear mixing model (LMM) to estimate representation coefficients and can be broadly categorized into binary and unified representations. In binary representation, each testing pixel is modeled separately as background and target using distinct dictionaries, and detection is performed by comparing their respective representation residuals. Representative binary detectors include the sparse representation-based detector (STD) \cite{chen2011sparse}, sparse representation-based binary hypothesis (SRBBH) \cite{zhang2014sparse}, combined sparse and collaborative representation (CSCR) \cite{li2015combined}, target dictionary construction-based STD (TDC-STD) \cite{zhu2019target}, and weighted discriminative collaborative competitive representation (WDCCR) \cite{liu2024weighted}. In contrast, unified representation models all pixels jointly using a combined background-target dictionary, with specific regularization priors imposed on the coefficients to separate background and target components. Typical unified detectors include the decomposition model with background dictionary learning (DM-BDL) \cite{cheng2021decomposition}, dual sparse constraints (DSC) \cite{shen2022dual}, double dictionary-based nonlinear representation for target detection (DDNRTD) \cite{wang2022double}, and the learning single spectral abundance (LSSA) method \cite{zhu2023learning}. These unified approaches excel at subpixel target detection by effectively separating the target from the background. To further preserve the spatial-spectral structure of HSI and capture intrinsic correlations, tensor decomposition-based methods have been developed. Zhao \emph{et al.} \cite{zhao2022hyper} proposed a method based on nonlocal similarity and rank-1 tensor decomposition, which effectively extracts spatial-spectral features and achieves robust performance under few-shot detection scenarios. Kong \emph{et al.} \cite{kong2025hyper} introduced a tensor approximation model employing a tensor double nuclear norm to capture both global and local low-rank properties, as well as the nonlocal similarity of the background. Although these methods offer physical and mathematical interpretability along with promising generalization capabilities, they face challenges in practical applications. On the one hand, handcrafted regularizations may fail to accurately capture the intrinsic characteristics of specific datasets. On the other hand, limited nonlinear representation capacity hampers their ability to fully discriminate between background and target.
\subsubsection{Deep Learning-Based Methods}
These methods leverage neural networks to extract discriminative features or patterns from data for HTD. However, the scarcity of labeled samples poses significant challenges in training effective and generalizable models. To address this issue, transfer learning, sample augmentation, and unsupervised learning strategies have been explored. Transfer learning aims to utilize knowledge from a source domain with abundant labeled samples and transfer it to the target domain. For example, the convolutional neural network-based detector (CNND) \cite{li2017transferred} and the meta-learning method with a Siamese network (MLSN) \cite{wang2022meta} pretrain on classification datasets and transfer knowledge to detection datasets. Beyond spectral knowledge transfer, a recent method, transfer learning from high-resolution RGB images for HTD (TLH$^2$TD) \cite{wu2024transfer}, incorporates spatial features from RGB images containing similar small targets, employing box-wise detection to capture morphological characteristics. While these transfer learning methods alleviate the issue of limited labeled data, they may suffer from domain drift, where the knowledge acquired from the source domain becomes misaligned when applied to the target domain. Sample augmentation methods expand the training dataset by generating synthetic samples and train supervised binary classification models for detection. For instance, HTD-Net \cite{zhang2020htd} employs a U-shaped autoencoder to generate target pixels and uses a linear strategy to select background pixels. The two-stream CNN for target detection (TSCNTD) \cite{zhu2020two} identifies background pixels via sparse representation and mixes them with target priors to ensure sufficient target samples. The Siamese Transformer network for target detection (STTD) \cite{rao2022siamese} extracts high-purity background samples through endmember extraction, unmixing, and refiltering, while mixed target and background samples are generated via random linear mixing with target spectra. The triplet spectral-wise Transformer-based target detector (TSTTD) \cite{jiao2023triplet} treats original hyperspectral pixels as background and synthesizes target samples by mixing them with the target spectrum. Despite their promising performance, sample augmentation methods may introduce knowledge bias and noise, leading to distribution mismatches between training and testing data, which can cause missed detections or false alarms. Unsupervised methods typically employ autoencoders to reconstruct HSI, aiming to extract critical latent features, enhance relevant information, and suppress interference for improved background-target discrimination. For example, the distance-constrained stacked sparse autoencoder (DCSSAE) \cite{shi2019discriminative} maximizes the distance between the target prior and typical background spectra to extract discriminative features. The unconstrained linear mixture model with deep learning (ULMMDL) \cite{li2022target} uses a hierarchical denoising autoencoder to enhance target spectral consistency and suppress background interference, followed by a two-step subspace projection for target enhancement. Given the availability of sufficient background pixels, reconstructing the background has emerged as an effective detection strategy. The target suppression constraint-based background learning (BLTSC) method \cite{xie2020background} employs an AAE to estimate the background distribution, detecting targets by measuring the spectral distance between the reconstructed background and the original HSI. Building on this, the orthogonal subspace-guided variational autoencoder (OS-VAE) \cite{tian2024hyperspectral} introduces an orthogonal subspace loss to reduce the impact of suspicious target samples, thereby enhancing background purity. Recently, contrastive learning, a prominent self-supervised approach, has gained attention for its ability to extract supervisory information from unlabeled data and capture instance distinctions. The self-supervised spectral-level contrastive learning-based HTD (SCLHTD) method \cite{wang2023self} partitions spectral bands into even and odd groups, processing them through two AAEs to extract intrinsic feature representations. Spectral-level contrastive training enforces semantic consistency among feature vectors of the same pixel while distinguishing them from those of others. Additionally, the spectral graph contrast clustering assignment (SGCCA) method \cite{chen2024target} integrates graph theory by constructing a spectral graph of pixel spectra and leveraging graph convolution networks and Transformers to extract local-global structural features, providing new insights for HTD. While these unsupervised methods eliminate the need for large labeled samples and demonstrate promising performance, their reliance on spectral feature comparison while often overlooking distribution learning, may limit robustness when applied to complex real-world scenarios.
\subsection{Diffusion Models}
Diffusion models have garnered significant attention due to their ability to learn data distributions and generate high-quality, realistic samples. The pioneering work, DDPM \cite{NEURIPS2020_4c5bcfec}, captures data distributions by gradually adding noise to the data to form a noisy sequence, followed by a reverse denoising process to recover the original data. The training objective focuses on maximizing the variational lower bound. Compared with AEs and GANs, DDPM demonstrates superior generation quality and robustness. As a result, DDPM-based models have been widely applied across various fields. In HSI processing, diffusion models have primarily been explored for tasks such as HSI classification \cite{chen2023spectraldiff, zhou2024exploring}, unmixing \cite{deng2024diff, zeng2024unmix}, and super-resolution \cite{wu2023hsr, liu2024aspectral}. However, their application in HTD remains limited. Ma \emph{et al.} \cite{ma2023bsdm} introduced the first diffusion model for hyperspectral anomaly detection, treating the background as Gaussian noise and constructing a specialized Markov chain to suppress it. To better estimate clean backgrounds, Chen \emph{et al.} \cite{Chen2025dwsdiff} proposed a dual-window spectral diffusion model that progressively mixes each anomaly pixel with surrounding background information in the forward process and restores the clean background in the reverse process. Li \emph{et al.} \cite{Li2024ajoint} designed a conditional diffusion model for underwater target detection, generating multiple underwater images to simulate spectral variability, guided by principal component-based HSI. Despite these advancements, these methods primarily use diffusion models as auxiliary tools to enhance background-target separation. A fully end-to-end framework based on diffusion models for HTD remains unexplored.

In this paper, we propose a spectral conditional diffusion model that directly generates the detection map in an end-to-end manner. Specifically, the model treats background and target confidence scores as negative and positive samples, respectively, and diffuses them into noisy sequences during the forward process. A lightweight denoising network is designed to reconstruct clean confidence scores, conditioned on the corresponding spectra, and guided by a tailored background-target separation loss that enhances discriminability.  In the reverse process, the model starts from Gaussian noise and progressively generates the detection map, conditioned on the original HSI. By integrating distribution learning into a fully end-to-end framework, the proposed model significantly improves both detection quality and efficiency.
\begin{figure*}[!t]
	\centering
	\includegraphics[width= 1.0 \textwidth]{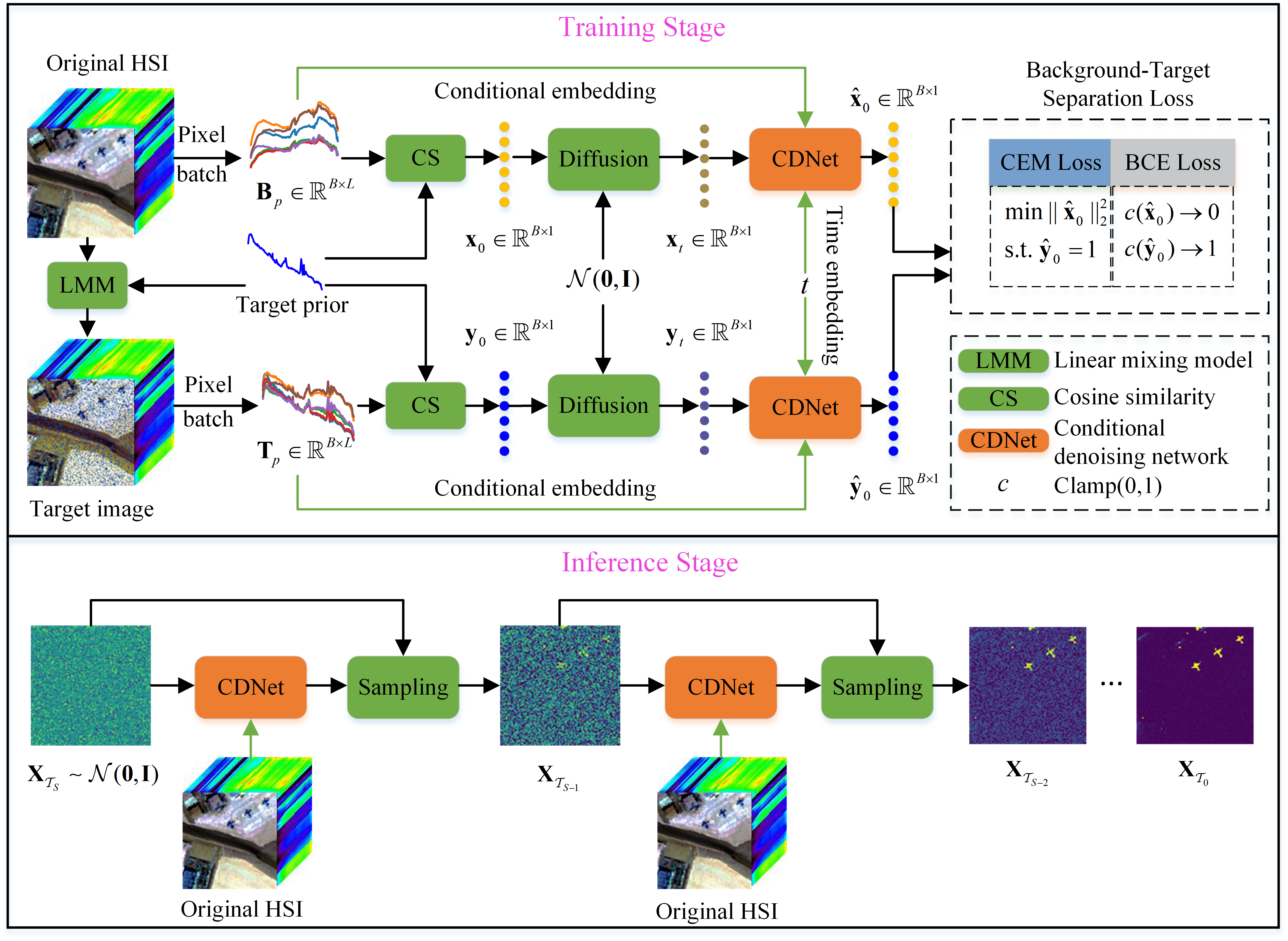}
	\caption{Overall framework of the proposed SCDM-HTD.}
	\label{framework}
\end{figure*}
\section{Proposed Method}
\label{method}
\subsection{Problem Formulation}
Given an HSI ${\bf Z} \in \mathbb{R}^{N \times L}$ with $N$ pixels and $L$ spectral bands,  and a known target spectrum ${\bf d} \in \mathbb{R}^{L}$, the goal of HTD is to classify each pixel in ${\bf Z}$ as either target or background. For the $i$-th pixel ${\bf Z}^i$, the detection result ${\bf F}^i$ is determined by
\begin{equation}
	{\bf F}^i = {\rm D}_z({\bf Z}^i, {\bf d})
\end{equation}
where ${\rm D}_z$ is a detection function that computes the confidence score indicating how likely a testing pixel matches the target spectrum. The common strategy for HTD typically focuses on comparing spectral features, which heavily rely on the effectiveness of feature extraction and may be sensitive to spectral variation. In this paper, we reformulate the detection task as a generation problem, employing a diffusion model to capture the distributional discrepancy between the background and target, ultimately generating an ideal confidence map. 
\subsection{Overall Framework}
HTD can be formulated as the task of generating confidence scores conditioned on pixel spectra. Inspired by the high-quality data generation capabilities of diffusion models, we propose SCDM-HTD to capture the confidence score distribution and iteratively refine the desired detection map. The overall framework of SCDM-HTD is illustrated in Fig. \ref{framework}.

The training stage consists of three key steps: confidence score construction, confidence score diffusion, and spectral conditional denoising. In the first step, the original hyperspectral pixel is treated as the background, while a mixed pixel, obtained by mixing it with the target prior via the LMM, represents the target. A batch of randomly sampled target and background pixels is compared with the target prior, and their cosine similarities define the positive and negative confidence scores. During diffusion, Gaussian noise is progressively added to the confidence score pairs, forming noisy sequences. These sequences are then denoised using a CDNet, where the corresponding pixel spectra serve as guiding conditions to capture the distribution of confidence scores. To effectively recover confidence scores and enhance target-background separability, the denoising process is optimized with a background-target separation loss. This loss integrates CEM loss, which enhances target responses while suppressing the background, and BCE loss, which enforces binary classification of background and target pixels for improved separation.

During inference, the well-trained denoising network takes the original HSI as conditional guidance. Starting from Gaussian noise, the confidence map is iteratively refined through reverse denoising and sampling, ultimately producing the final detection result.
\subsection{Confidence Score Construction}
Since prior confidence scores are unavailable, we employ a data augmentation strategy based on hyperspectral imaging to generate a sufficient number of positive and negative confidence scores.

Hyperspectral imaging follows the radiance transfer model:
\begin{equation}
	L_{\rm SR}(\lambda)=L_{\rm OGR}(\lambda)R(\lambda) + L_{\rm SUR}(\lambda)R_{\rm SUR}(\lambda)  + L_{\rm AS}(\lambda)
\end{equation}
where $L_{\rm SR}(\lambda)$ denotes the spectral radiance of the observed pixel at wavelength $\lambda$, $L_{\rm OGR}(\lambda)$ represents the overall ground-reflected radiance of the observed object, and $R(\lambda)$ is its reflectance. Additionally, $L_{\rm SUR}(\lambda)$ and $R_{\rm SUR}(\lambda)$ correspond to the radiance and reflectance of the surrounding region, respectively, while $L_{\rm AS}(\lambda)$ accounts for atmospheric scattering. In summary, the radiance of an observed pixel comprises three primary components: object reflection, background reflection, and atmospheric scattering. 

Since atmospheric scattering has a smaller effect, we focus on object reflection and background reflection to simulate the imaging process. Specifically, we treat the original hyperspectral pixels as background samples and generate target samples by randomly mixing them with the target prior:
\begin{align}
	{\bf B}^i &= {\bf Z}^i, \\
	{\bf T}^i &= \gamma{\bf B}^i + (1-\gamma) {\bf d}
\end{align}
where ${\bf B} \in \mathbb{R}^{N \times L}$ denotes the background samples, ${\bf T} \in \mathbb{R}^{N \times L}$ denotes the generated target samples, and $\gamma$ is a coefficient randomly selected from [0, 0.4] to ensure that the target material contributes more significantly to the mixture. This strategy effectively simulates subpixel targets while maintaining a diverse and balanced sample set for training.

To obtain confidence scores for diffusion, the background and target pixels are compared separately with the target prior using cosine similarity:
\begin{align}
	{\bf X}^i &= {\bf d}^T{\bf B}^i / (\|{\bf B}^i\|_2\|{\bf d}\|_2), \\
	{\bf Y}^i &= {\bf d}^T{\bf T}^i / (\|{\bf T}^i\|_2\|{\bf d}\|_2)
\end{align}
where $\|\cdot\|_2$ denotes the L2 norm. The resulting sets $\{{\bf X}^i\}_{i=1}^N$, and $\{{\bf Y}^i\}_{i=1}^N$ correspond to the negative and positive confidence scores, respectively.
\begin{figure}[!t]
	\centering
	\includegraphics[width= 0.48 \textwidth]{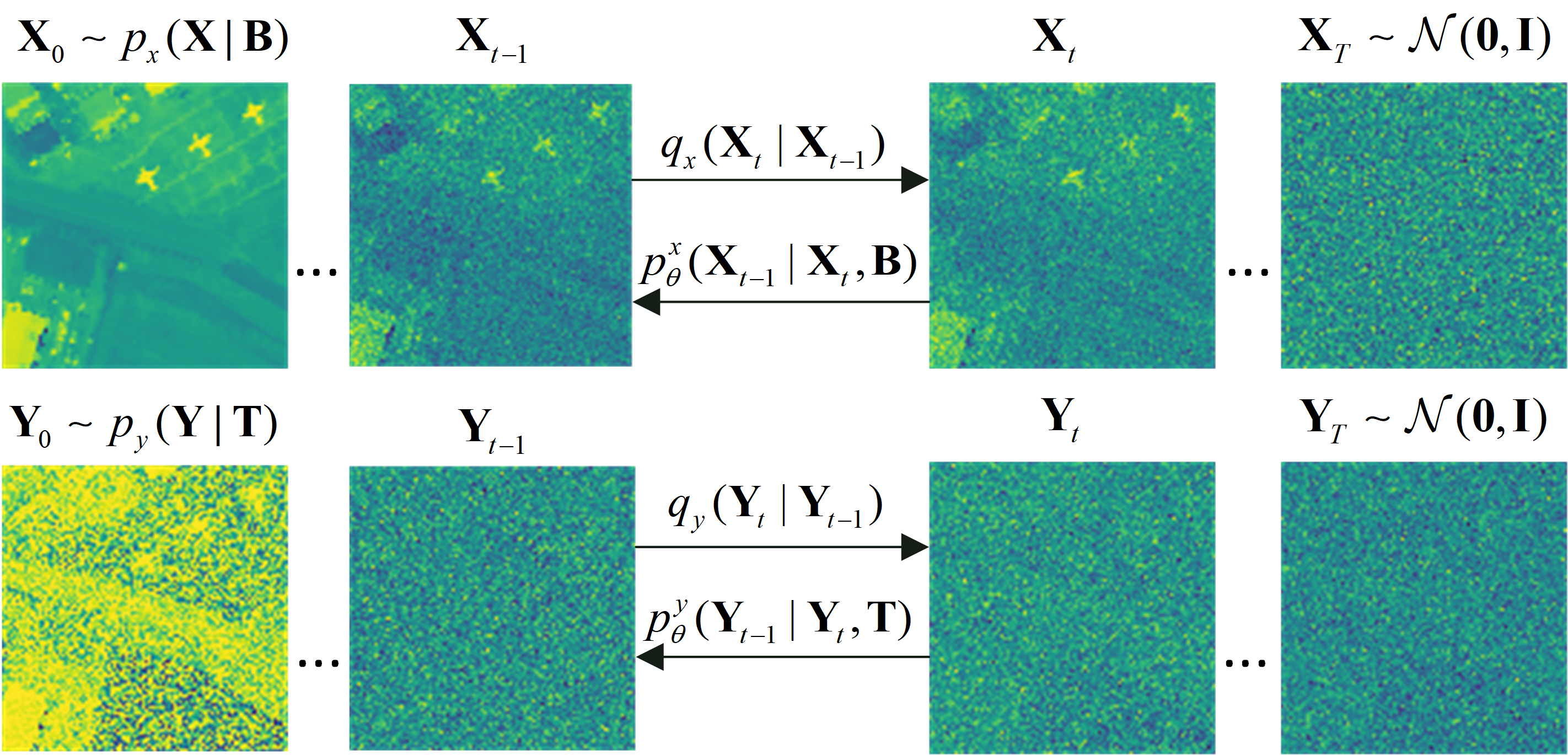}
	\caption{Forward and reverse processes of SCDM-HTD: The forward diffusion progressively adds Gaussian noise to the binary confidence maps, gradually corrupting them. In contrast, the reverse sampling process iteratively denoises the maps to recover the confidence scores, conditioned on the corresponding pixel spectra.}
	\label{markovs}
\end{figure}
\subsection{Confidence Score Diffusion}
To capture the distributional discrepancy between target and background, we propose a binary score diffusion model that diffuses the confidence scores associated with both positive (target) and negative (background) samples, and reconstructs them conditioned on the spectral characteristics of each pixel. Drawing inspiration from DDPM, this model comprises a forward diffusion process and a reverse sampling process, as illustrated in Fig. \ref{markovs}. In the forward process, Gaussian noise is progressively added to the confidence scores, leading to their corruption. Conversely, the reverse process begins with Gaussian noise and progressively recovers the scores, conditioned on the corresponding pixel spectra.
\subsubsection{Forward Diffussion Process}
Let a random batch of background and target pixels be denoted as ${\bf B}_p \in \mathbb{R}^{B \times L}$ and ${\bf T}_p \in \mathbb{R}^{B \times L}$, where $B$ is the batch size. The corresponding confidence scores are represented as ${\bf x}_0 \in \mathbb{R}^{B \times 1}$ and ${\bf y}_0 \in \mathbb{R}^{B \times 1}$, which are assumed to follow the desired data distributions: ${\bf x}_0 \sim q_x({\bf x}_0)$ and ${\bf y}_0 \sim q_y({\bf y}_0)$. The forward diffusion processes, $q_x({\bf x}_t|{\bf x}_{t-1})$ and $q_y({\bf y}_t|{\bf y}_{t-1})$, define a shared Markov chain that generates latent variables $\{{\bf x}_t\}_{t=1}^T$ and $\{{\bf y}_t\}_{t=1}^T$ by gradually adding Gaussian noise at each timestep $t$:
\begin{align}
q_x({\bf x}_{1:T}|{\bf x}_0)&=\prod_{t=1}^{T}q_x({\bf x}_t|{\bf x}_{t-1}), \\
q_x({\bf x}_t|{\bf x}_{t-1}) &= \mathcal{N}({\bf x}_t;\sqrt{\alpha_t}{\bf x}_{t-1}, (1-\alpha_t){\bf I})
\end{align}
where $\{\alpha_t\}_{t=1}^T$ are scalar parameters related to the noise variance, and ${\bf I}$ denotes the identity matrix. For notational simplicity, the contrastive part of ${\bf y}$ is omitted in the following derivations. The noisy confidence scores ${\bf x}_T$ and ${\bf y}_T$ at a large timestep $T$ approach standard Gaussian noise. Using the reparameterization trick, we can sample from $q_x({\bf x}_t|{\bf x}_0)$ and $q_y({\bf y}_t|{\bf y}_0)$ in closed form:
\begin{align}
q_x({\bf x}_t|{\bf x}_0) &= \mathcal{N}({\bf x}_t;\sqrt{\bar{\alpha}_t}{\bf x}_0, (1-\bar{\alpha}_t){\bf I}) 
\end{align}
where $\bar{\alpha}_t=\prod_{i=1}^{t}\alpha_i$.
\subsubsection{Reverse Sampling Process} Since the true reverse distributions $q_x({\bf x}_{t-1}|{\bf x}_t)$ and $q_y({\bf y}_{t-1}|{\bf y}_t)$ are intractable, we approximate them using the CDNet $f_\theta$, where $\theta$ represents the network parameters. The reverse sampling process is defined as a Markov chain comprising the parameterized conditional Gaussian transitions $p_\theta^x({\bf x}_{t-1}|{\bf x}_t, {\bf B}_p)$ and $p_\theta^y({\bf y}_{t-1}|{\bf y}_t,  {\bf T}_p)$:
\begin{align}
p_\theta^x({\bf x}_{0:T}|{\bf B}_p)&=p_x({\bf x}_T)\prod_{t=1}^{T}p_\theta^x({\bf x}_{t-1}|{\bf x}_{t}, {\bf B}_p), \\
p_\theta^x({\bf x}_{t-1}|{\bf x}_{t},{\bf B}_p) &= \mathcal{N}({\bf x}_{t-1};\mu_\theta({\bf x}_{t},{\bf B}_p, t), \sigma_t^2{\bf I})	
\end{align}
where $p_x({\bf x}_T)=\mathcal{N}({\bf x}_T, {\bf 0}, {\bf I})$, $\mu_\theta$ is the parameterized mean, and $\sigma_t^2$ is the corresponding variance. To maximize the variational lower bound, we approximate $q_x({\bf x}_{t-1}|{\bf x}_t, {\bf x}_0,{\bf B}_p)$ with $p_\theta^x({\bf x}_{t-1}|{\bf x}_t, {\bf B}_p)$. The mean $\mu_\theta$ and variance $\sigma_t^2$ are given by
\begin{align}
	\mu_\theta({\bf x}_{t},{\bf B}_p, t)&=\frac{\sqrt{\bar{\alpha}_{t-1}}(1-\alpha_t)}{1-\bar{\alpha}_t}{\hat{\bf x}}_0+\frac{\sqrt{\alpha_t}(1-\bar{\alpha}_{t-1})}{1-\bar{\alpha}_t}{\bf x}_t,\\
	\sigma_t^2 &= \frac{(1-\bar{\alpha}_{t-1})(1-\alpha_t)}{1-\bar{\alpha}_t}
	\label{posterior}
\end{align}
where ${\hat{\bf x}}_0=f_\theta({\bf x}_{t},{\bf B}_p, t)$ is the denoised estimate of the initial confidence score. Once the CDNet $f_\theta$ is well-trained, we can perform HTD by iteratively sampling from Gaussian noise, guided by the original HSI:
\begin{equation}
	{\bf X}_{t-1}= \mu_\theta({\bf X}_{t},{\bf Z}, t) + \sigma_t\epsilon
\end{equation}
where $\epsilon\sim\mathcal{N}({\bf 0}, {\bf I})$ and $t \in \{T,\dots,1\}$. Note that $\sigma_t$ is experimentally set to $\sqrt{1-\alpha_t}$ for simplicity and consistency with the forward process.
\begin{figure}[!t]
	\centering
	\includegraphics[width= 0.5 \textwidth]{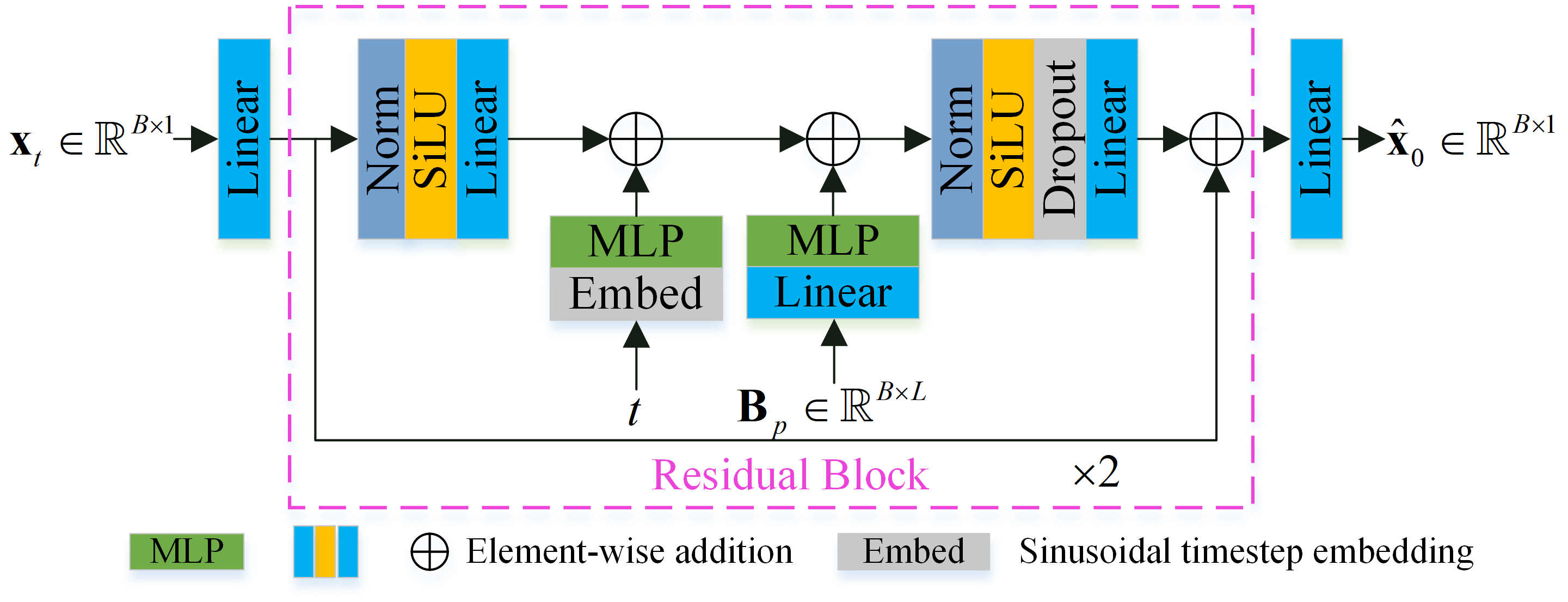}
	\caption{Flowchart of CDNet.}
	\label{CD_Net}
\end{figure}
\subsection{Spectral Conditional Denoising}
To effectively recover confidence scores and distinguish between background and target, spectral conditional denoising is applied, involving the design of the denoising network and the formulation of the objective loss function.
\begin{table}[!t]  
	\centering  
	\caption{Detailed configuration of CDNet}  
	\footnotesize  
	\tabcolsep = 0.8pt  
	\renewcommand\arraystretch{1.3}  
	\begin{tabular}{c|c|c|c|c}  
		\hline  
		\multicolumn{2}{c|}{Layer} & Group size & Input size & Output size \\  
		\hline  
		Initial layer & Linear & - & $B\times 1$ & $B\times 128$ \\
		\hline
		Time embedding & Embed & - & $B\times 1$ & $B\times 128$ \\
		\hline
		Conditional embedding & Linear & - & $B\times L$ & $B\times 128$ \\
		\hline  
		\multirow{2}*{MLP} &Linear+SiLU&-
		& $B\times 128$ & $B\times 512$ \\
		&Linear&-
		& $B\times 512$ & $B\times 512$ \\
		\hline  
		\multirow{4}*{Residual block} & GroupNorm+SiLU&32
		 & $B\times 128$ & $B\times 128$ \\  
		& Linear & - & $B\times 128$ & $B\times 512$ \\  
		&GroupNorm+SiLU&32
		& $B\times 512$ & $B\times 512$ \\
		&Dropout+Linear&-
		& $B\times 512$ & $B\times 128$ \\
		\hline
		Final layer & Linear & - & $B\times 128$ & $B\times 1$ \\
		\hline
		% Repeat for more rows  
	\end{tabular}  
	\label{CD_Net_config}  
\end{table}
\subsubsection{Conditional Denoising Network}
The classical DDPM employs a UNet architecture with attention mechanisms as its denoising network. However, this architecture is not well-suited for denoising confidence scores due to the scalar nature of the data and the high computational efficiency demands in practical applications. To address this, we propose CDNet, a lightweight network based on a fully connected residual architecture, as shown in Fig. \ref{CD_Net}. The CDNet $f_\theta({\bf x}_t, {\bf B}_p, t)$ consists of an initial linear layer, intermediate residual layers, and a final linear layer. The initial linear layer maps each confidence score to a higher-dimensional vector space. The intermediate residual layers comprise two residual blocks, where each block first extracts nonlinear features and then fuses them with the timestep and conditional embeddings. Similar to DDPM, the timestep is embedded using the Transformer sinusoidal position embedding \cite{NIPS2017_3f5ee243}, followed by an MLP. For the conditional embedding, the corresponding pixel spectra of the confidence scores are injected through a simple linear layer and an MLP to guide the distribution learning of confidence scores. The fused features then pass through an additional MLP block to integrate and refine the features. To prevent gradient vanishing, the refined features are added to the initial features via a residual connection. Finally, the last linear layer recovers the extracted features into confidence scores. Table \ref{CD_Net_config} provides the detailed configuration of CDNet $f_\theta({\bf x}_t, {\bf B}_p, t)$.
\subsubsection{Objective Loss Function} The objective of the 
conditional diffusion model is to maximize the evidence lower bound (ELBO), which acts as a surrogate for maximizing the log-likelihood of the data. The ELBO is given by:
\begin{equation}
	{\rm ELBO}=\mathbb{E}_{q_x({\bf x}_{1:T}|{\bf x}_0)}{\rm log}\frac{p_\theta^x({\bf x}_{0:T}|{\bf B}_p)}{q_x({\bf x}_{1:T}|{\bf x}_0)}.
\end{equation}
According to the Bayes' theorem, the ELBO can be rewritten as 
\begin{align}
	{\rm ELBO}=&-{\rm D}_{\rm KL}(q_x({\bf x}_T|{\bf x}_0)\|p_x({\bf x}_T)) \notag\\&-\sum_{t=2}^{T}{\rm D}_{\rm KL}(q_x({\bf x}_{t-1}|{\bf x}_t,{\bf x}_0)\|p_\theta^x({\bf x}_{t-1}|{\bf x}_{t},{\bf B}_p
	))\notag\\&+\mathbb{E}_{q_x({\bf x}_{1:T}|{\bf x}_0)}{\rm log}p_\theta^x({\bf x}_0|{\bf x}_1, {\bf B}_p)
\end{align}
where ${\rm D}_{\rm KL}$ denotes the Kullback-Leibler (KL) divergence. The first term represents the prior matching term, the second term represents the posterior matching term, and the last term represents the reconstruction term. Since the prior matching term does not contain any learnable parameters, we focus on optimizing the last two terms, which leads to the diffusion loss:
\begin{equation}
	\mathcal{L}_{\rm diff}=\frac{1}{B}\left(\|{\bf x}_0-f_\theta({\bf x}_{t},{\bf B}_p, t)\|_2^2+\|{\bf y}_0-f_\theta({\bf y}_{t},{\bf T}_p, t)\|_2^2\right).
	\label{diff}
\end{equation}
The diffusion loss ensures that the reconstructed confidence scores approximate the original ones, aiding in the recovery of the original confidence scores.  However, this reconstruction objective alone does not sufficiently distinguish between target and background. Additionally, since not all augmented confidence scores are accurate, the denoising process remains weakly supervised. To address these issues in the HTD task, we propose a background-target separation loss, which combines CEM loss and BCE loss to more effectively differentiate between background and target.

Inspired by the classical CEM method, the CEM loss aims to enhance the target response while suppressing the background by maximizing the target output and minimizing the background output energy. The corresponding optimization problem is given by:
\begin{equation}
	\min_\theta\|f_\theta({\bf x}_{t},{\bf B}_p, t)\|_2^2\;\;{\rm s.t.}f_\theta({\bf y}_{t},{\bf T}_p, t)=1.
\end{equation}
Here, the optimization term minimizes the background output energy, while the constraint ensures that the target response is exactly 1. By applying the Lagrange multiplier method, the CEM loss can be formulated as follows:
\begin{equation}
	\mathcal{L}_{\rm CEM}=\frac{1}{B}\left(\|f_\theta({\bf x}_{t},{\bf B}_p, t)\|_2^2+\mu\|f_\theta({\bf y}_{t},{\bf T}_p, t)-1\|_1\right)
\end{equation}
where $\mu$ is a penalty parameter. Compared to the diffusion loss in \eqref{diff}, the CEM loss does not aim to fully recover the confidence scores. Instead, it focuses on enhancing the target confidence scores while suppressing the background confidence scores, effectively separating them.

To further enhance detection robustness, pixelwise classification can be incorporated. Therefore, we introduce the BCE loss to supervise the denoising process. By assigning the target and background confidence scores with label 1 and 0, respectively, the BCE loss can be formulated as
\begin{equation}
	\mathcal{L}_{\rm BCE}=-\frac{1}{2B}\sum_{i=1}^{B}{\rm log}\left(c\left(\hat{{\bf y}}_0^i\right)+g\right)+{\rm log}\left(c\left(1-\hat{{\bf x}}_0^i\right)+g\right)
\end{equation}
where ${\hat{\bf x}}_0=f_\theta({\bf x}_{t},{\bf B}_p, t)$, ${\hat{\bf y}}_0=f_\theta({\bf y}_{t},{\bf T}_p, t)$, $c$ is the clamp function that forces values into the interval $[0,1]$, and $g$  is a small positive value added to avoid undefined logarithms at zero. Each batch contains $B$ background confidence scores and $B$ corresponding target confidence scores. The BCE loss categories them into the negative and positive classes for separation.

In summary, the proposed background-target separation loss combines the CEM loss and the BCE loss as follows:
\begin{equation}
	\mathcal{L}=\mathcal{L}_{\rm CEM} + \lambda\mathcal{L}_{\rm BCE}
\end{equation}
where $\lambda$ is a regularization parameter that balances the contributions of these two loss terms.
\subsection{Reverse Sampling Acceleration}
The reverse sampling process in DDPM follows a Markov chain, which typically requires many iterations and does not meet the efficiency demands of practical detection tasks. To accelerate the sampling process, inspired by DDIM, we reformulate the reverse process as a non-Markovian chain. This reformulation enables faster sampling by skipping steps.

DDIM ignores the transition between adjacent probability distributions and focuses instead on the closed-form transformation from the initial state to the noisy state. The reverse sampling process is formulated as:
\begin{equation}
	{\bf X}_{t-1}=\sqrt{\bar{\alpha}_{t-1}}{\hat{\bf X}}_0+ \sqrt{1-\bar{\alpha}_{t-1}-\sigma_{t}^2}\epsilon_\theta({\bf X}_{t}, {\bf B},t)+\sigma_{t}\epsilon_{t}
\end{equation}
where ${\hat{\bf X}}_0=f_\theta({\bf X}_{t},{\bf B}, t)$, $\sigma_t^2=\eta\frac{1-\bar{\alpha}_{t-1}}{1-\bar{\alpha}_t}\beta_t$, and $\eta$ is a coefficient. The sub-sequence $t\in[\tau_S,\dots,\tau_1]$ represents a segment of the sequence $[T,\dots,1]$ of length $S$. When $\eta=0$, the reverse sampling process becomes deterministic with no added random noise:
\begin{equation}
	{\bf X}_{t-1}=\sqrt{\bar{\alpha}_{t-1}}{\hat{\bf X}}_0+ \sqrt{1-\bar{\alpha}_{t-1}}\epsilon_\theta({\bf X}_{t}, {\bf B},t).
\end{equation}

In this work, we focus on estimating the initial confidence scores rather than the noise. Thus, the reverse sampling can be expressed as
\begin{equation}
	{\bf X}_{t-1}=\sqrt{\bar{\alpha}_{t-1}}{\hat{\bf X}}_0+ \sqrt{1-\bar{\alpha}_{t-1}}\frac{{\bf X}_{t}-\sqrt{\bar{\alpha}_{t}}{\hat{\bf X}}_0}{(1-\sqrt{\bar{\alpha}_{t}})}.
\end{equation}
This accelerated sampling process transforms Gaussian noise into a deterministic detection map in just a few steps, significantly improving efficiency for practical applications.
%\frac{{\bf X}_{\tau_{S}}-\sqrt{1-\bar{\alpha}_{\tau_{S}}}\epsilon_\theta({\bf X}_{\tau_{S}}, {\bf B},\tau_{S} )}{\sqrt{\bar{\alpha}_{\tau_{S}}}}
%Since a prior confidence map ${\bf F}$ is unavailable, we introduce pseudo-background and pseudo-target confidence maps, thereby constructing a binary diffusion process. As shown in Fig. \ref{markovs}, the cosine similarity map between the target spectrum and the original HSI is used as the pseudo-background confidence map ${\bf X}$, while the cosine similarity map between the target spectrum and the mixed HSI (generated via linear mixing with the target spectrum) serves as the pseudo-target confidence map ${\bf Y}$.  This results in the need for a binary forward process that progressively perturbs ${\bf X}$ and ${\bf Y}$ to noise, where paired intermediate maps, i.e., (${\bf X}_1$, ${\bf Y}_1$),$\dots$, (${\bf X}_{T-1}$, ${\bf Y}_{T-1}$), are generated according to a Markov chain with fixed transition probabilities $q({\bf X}_{t}|{\bf X}_{t-1})$ and $q({\bf Y}_{t}|{\bf Y}_{t-1})$. The conditional distributions $p_\theta({\bf X}_{t-1}|{\bf X}_{t}, {\bf B})$ and $p_\theta({\bf Y}_{t-1}|{\bf Y}_{t}, {\bf T})$ are learned using the CDNet $f_\theta$, where ${\bf B}$ and ${\bf T}$ denote the original HSI and the mixed HSI, respectively. In the reverse process, noise is progressively reduced with $p_\theta({\bf X}_{t-1}|{\bf X}_{t}, {\bf B})$ to generate the desired detection map.
\section{Experiments}
\label{experiments}
\subsection{Datasets}
Experiments were conducted on four publicly available datasets to evaluate the effectiveness of the proposed method.
%\subsubsection{San Diego} This dataset was captured by the Airborne Visible/Infrared Imaging Spectrometer (AVRIS) over the San Diego airport, CA, USA. It contains $100\times 100$ pixels with a spatial resolution of 3.5 m, and each pixel consists of 189 bands after removing the low-quality and water vapor absorption bands (1-6, 33-35, 97, 107-113, 153-166, and 221-224). The targets to be detected are three airplanes, which together consist of 58 pixels.
\subsubsection{MUUL Gulfport
} This dataset was acquired by the Compact Airborne Spectrographic Imager (CASI-1500) over the Gulf Park Campus of the University of Southern Mississippi \cite{gader2013muufl}. The scence consists of $325\times 337$ pixels, with 64 spectral bands retained after removing noisy bands (1–4 and 69–72). Following the method in \cite{jiao2023triplet}, a $30 \times 30$ pixel subimage is extracted from the original HSI for experimentation. Two solid brown cloth panels within this subimage, comprising 10 pixels, are selected as the targets.
\subsubsection{HYDICE} This dataset was collected by the Hyperspectral Digital Imagery Collection Experiment (HYDICE) sensor \cite{mitchell1995hyperspectral}. The scene covers an urban area in California, USA, which contains $307 \times 307$ pixels with a spatial resolution of 1 m. After removing water vapor absorption and noisy bands (1-4, 76, 87, 101-111, 136-153, and 198-210), 162 bands remain. For detection experiments, a commonly used $80 \times 100$ pixel subimage is extracted. The targets in this subimage are several vehicles, occupying a total of 19 pixels.
\subsubsection{Airport-4} This dataset is part of the airport-beach-urban (ABU) dataset \cite{kang2017hyperspectral}, captured by the the Airborne Visible/Infrared Imaging Spectrometer (AVRIS). It contains $100\times 100$ pixels with a spatial resolution of 3.4 m, and each pixel contains 191 spectral bands. The targets to be detected are three airplanes, occupying a total of 60 pixels.
\subsubsection{Urban-2} This dataset is also part of the ABU dataset, which was captured by the AVIRIS sensor. The scene covers an urban area and has $100 \times 100$ pixels with a spatial resolution of 17.2 m. Each pixel contains 207 spectral bands. The detection targets are several regularly arranged man-made objects, covering a total of 155 pixels.

Since the target prior must be known, it can be either selected from a spectral library (if available) or obtained by choosing a representative target pixel from the HSI. In this work, the reference target spectrum for the MUUL Gulfport dataset is provided through laboratory spectrometer measurements. For the other datasets, the target prior is determined by selecting the pixel closest to the averaged target spectrum.
\subsection{Compared Methods and Evaluation Metrics}
The proposed SCDM-HTD is compared with eight representative competing methods, including OSP \cite{chang2005orthogonal}, CEM \cite{farrand1997mapping}, CSCR\footnote[1]{\url{https://fdss.bit.edu.cn/pub/fsyxhyxtktz/yjdw/js/b153191.htm}} \cite{li2015combined}, WDCCR\footnote[2]{\url{https://github.com/liurongwhm/WDCCR}} \cite{liu2024weighted}, BLTSC\footnote[3]{\url{https://github.com/zhangxin-xd/BLTSC}} \cite{xie2020background}, HTD-Net\footnote[4]{\url{https://fdss.bit.edu.cn/pub/fsyxhyxtktz/yjdw/js/b153191.htm}} \cite{zhang2020htd}, TSTTD\footnote[5]{\url{https://github.com/shendb2022/TSTTD}} \cite{jiao2023triplet}, and OS-VAE\footnote[6]{\url{https://github.com/CX-He/OS-VAE}} \cite{tian2024hyperspectral}. Among them, OSP and CEM are statistical-based methods, CSCR and WDCCR are representation-based methods, and BLTSC, HTD-Net, TSTTD, and OS-VAE are deep learning-based methods.

Both qualitative and quantitative metrics are used to assess detection performance. The qualitative evaluation includes visualized detection maps, 3D receiver operating characteristic (ROC) analysis \cite{wang20063d}, and the background-target separability diagram. The detection map provides an intuitive visualization of detection effectiveness. The 3D ROC curve represents the relationship among detection probability ${\rm P_d}$, false alarm rate ${\rm P_f}$, and threshold $\tau$. Three derived 2D ROC curves of (${\rm P_f}$, ${\rm P_d}$), ($\tau$, ${\rm P_f}$), and ($\tau$, ${\rm P_d}$) are used to evaluate detection effectiveness, background suppression, and target detectability, respectively. The background-target separability diagram further assesses the degree of separation between background and target. For quantitative evaluation, three basic areas under the ROC curves (AUCs) are considered: ${\rm AUC}_{(\rm P_f, P_d)}$, ${\rm AUC}_{(\tau, {\rm P_f})}$, and ${\rm AUC}_{(\tau, {\rm P_d})}$. Additionally, two composite metrics \cite{chang2020effective}, ${\rm AUC}_{{\rm OA}}$ and ${\rm AUC}_{{\rm SNPR}}$, are used to assess overall detection quality.
\subsection{Implementation Specification}
For the competing methods, we run the publicly available source codes with their optimal hyper-parameters as reported in the original papers. The proposed SCDM-HTD is implemented in Python 3.10.14 using the PyTorch 2.1.1 framework and trained on an NVIDIA GeForce RTX 3090 GPU with 24 GB memory.  The batch size $B$ is set to 256, and the total number of timestep $T$ is 500. The penalty parameter $\mu$ in the CEM loss is is fixed at 10 for all datasets. The regulariztion parameter $\lambda$,  which balances the CEM loss and the BCE loss, is set to 1, $10^{-3}$, 1, and $10^{-4}$ for MULL Gulfport, HYDICE, Airport-4, and Urban-2, respectively. For optimization, we use the AdamW optimizer with a weight decay of  $10^{-4}$. The learning rate is initialized at $10^{-4}$, follows a warmup schedule for the first $10\%$ epochs, and then decreases following a cosine decay schedule. The model is trained for a total of 200 epochs. During inference, the number of sampling steps $S$ is set to 5 for all datsets.
\subsection{Performance Comparison}
\subsubsection{Qualitative Comparison} 
\begin{figure*}[!t]%加*的作用是跨栏（双栏和单栏latex的区别）
	\centering
	\footnotesize
	\vspace{-3pt}
	\subfigure{
		\begin{minipage}[c]{0.05\textwidth}
			\centering
			\text{(a)}
		\end{minipage}
		\hspace{-5pt}
		\begin{minipage}[c]{1.0\textwidth}
		\includegraphics[width=0.08\textwidth]{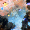}
		\includegraphics[width=0.08\textwidth]{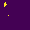} 
		\includegraphics[width=0.08\textwidth]{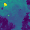} 
		\includegraphics[width=0.08\textwidth]{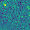} 
		\includegraphics[width=0.08\textwidth]{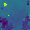} 
		\includegraphics[width=0.08\textwidth]{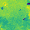}
		\includegraphics[width=0.08\textwidth]{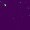}
		\includegraphics[width=0.08\textwidth]{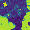} 
		\includegraphics[width=0.08\textwidth]{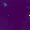} 
		\includegraphics[width=0.08\textwidth]{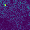} 
		\includegraphics[width=0.08\textwidth]{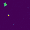} 
		\end{minipage}
	}\\
	\vspace{-3pt}
	\subfigure{
		\begin{minipage}[c]{0.05\textwidth}
			\centering
			\text{(b)}
		\end{minipage}
		\hspace{-5pt}
		\begin{minipage}[c]{1.0\textwidth}
			\includegraphics[width=0.08\textwidth]{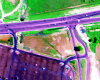}
			\includegraphics[width=0.08\textwidth]{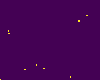} 
			\includegraphics[width=0.08\textwidth]{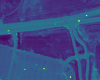} 
			\includegraphics[width=0.08\textwidth]{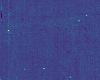} 
			\includegraphics[width=0.08\textwidth]{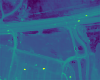} 
			\includegraphics[width=0.08\textwidth]{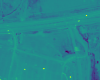}
			\includegraphics[width=0.08\textwidth]{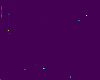}
			\includegraphics[width=0.08\textwidth]{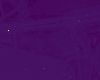}
			\includegraphics[width=0.08\textwidth]{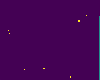} 
			\includegraphics[width=0.08\textwidth]{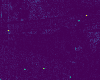} 
			\includegraphics[width=0.08\textwidth]{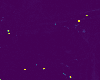} 
		\end{minipage}
	}\\
	\vspace{-3pt}
	\subfigure{
		\begin{minipage}[c]{0.05\textwidth}
			\centering
			\text{(c)}
		\end{minipage}
		\hspace{-5pt}
		\begin{minipage}[c]{1.0\textwidth}
				\includegraphics[width=0.08\textwidth]{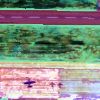}
				\includegraphics[width=0.08\textwidth]{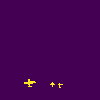} 
				\includegraphics[width=0.08\textwidth]{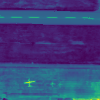} 
				\includegraphics[width=0.08\textwidth]{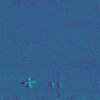} 
				\includegraphics[width=0.08\textwidth]{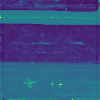} 
				\includegraphics[width=0.08\textwidth]{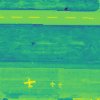}
				\includegraphics[width=0.08\textwidth]{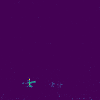}
				\includegraphics[width=0.08\textwidth]{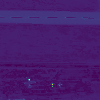}
				\includegraphics[width=0.08\textwidth]{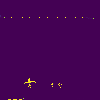} 
				\includegraphics[width=0.08\textwidth]{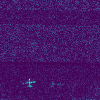} 
				\includegraphics[width=0.08\textwidth]{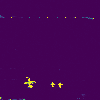} 
		\end{minipage}
	}\\
	\vspace{-3pt}
	\subfigure{
		\begin{minipage}[c]{0.05\textwidth}
			\centering
			\text{(d)}
		\end{minipage}
		\hspace{-5pt}
		\begin{minipage}[c]{1.0\textwidth}
			\includegraphics[width=0.08\textwidth]{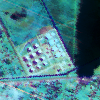}
			\includegraphics[width=0.08\textwidth]{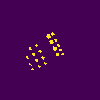} 
			\includegraphics[width=0.08\textwidth]{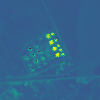} 
			\includegraphics[width=0.08\textwidth]{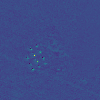} 
			\includegraphics[width=0.08\textwidth]{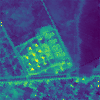} 
			\includegraphics[width=0.08\textwidth]{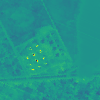}
			\includegraphics[width=0.08\textwidth]{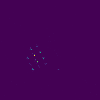}
			\includegraphics[width=0.08\textwidth]{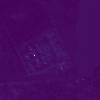}
			\includegraphics[width=0.08\textwidth]{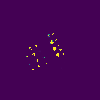} 
			\includegraphics[width=0.08\textwidth]{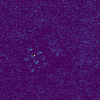} 
			\includegraphics[width=0.08\textwidth]{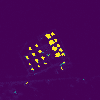}
		\end{minipage}
	}\\
	\vspace{-3pt}
	\subfigure{
		\begin{minipage}[c]{0.05\textwidth}
			\centering
			\text{}
		\end{minipage}
		\hspace{-5pt}
		\begin{minipage}[c]{1.0\textwidth}
			\parbox{0.08\textwidth}{\centering \text{False-color}}
			\parbox{0.08\textwidth}{\centering \text{Ground truth}}
			\parbox{0.08\textwidth}{\centering \text{OSP}}
			\parbox{0.08\textwidth}{\centering \text{CEM}}
			\parbox{0.08\textwidth}{\centering \text{CSCR}}
			\parbox{0.08\textwidth}{\centering \text{WDCCR}}
			\parbox{0.08\textwidth}{\centering \text{BLTSC}}
			\parbox{0.08\textwidth}{\centering \text{HTD-Net}}
			\parbox{0.08\textwidth}{\centering \text{TSTTD}}
			\parbox{0.08\textwidth}{\centering \text{OS-VAE}}
			\parbox{0.08\textwidth}{\centering \text{SCDM-HTD}}
		\end{minipage}
	} \caption{Visualized results of the competing methods on (a) MULL Gulfport, (b) HYDICE, (c) Airport-4, and (d) Urban-2.}
	\label{detection_results}
\end{figure*}
\begin{figure*}[htbp]
	\centering
	\begin{minipage}{1.0\textwidth}
		\subfigure{
			\includegraphics[width=0.25\textwidth]{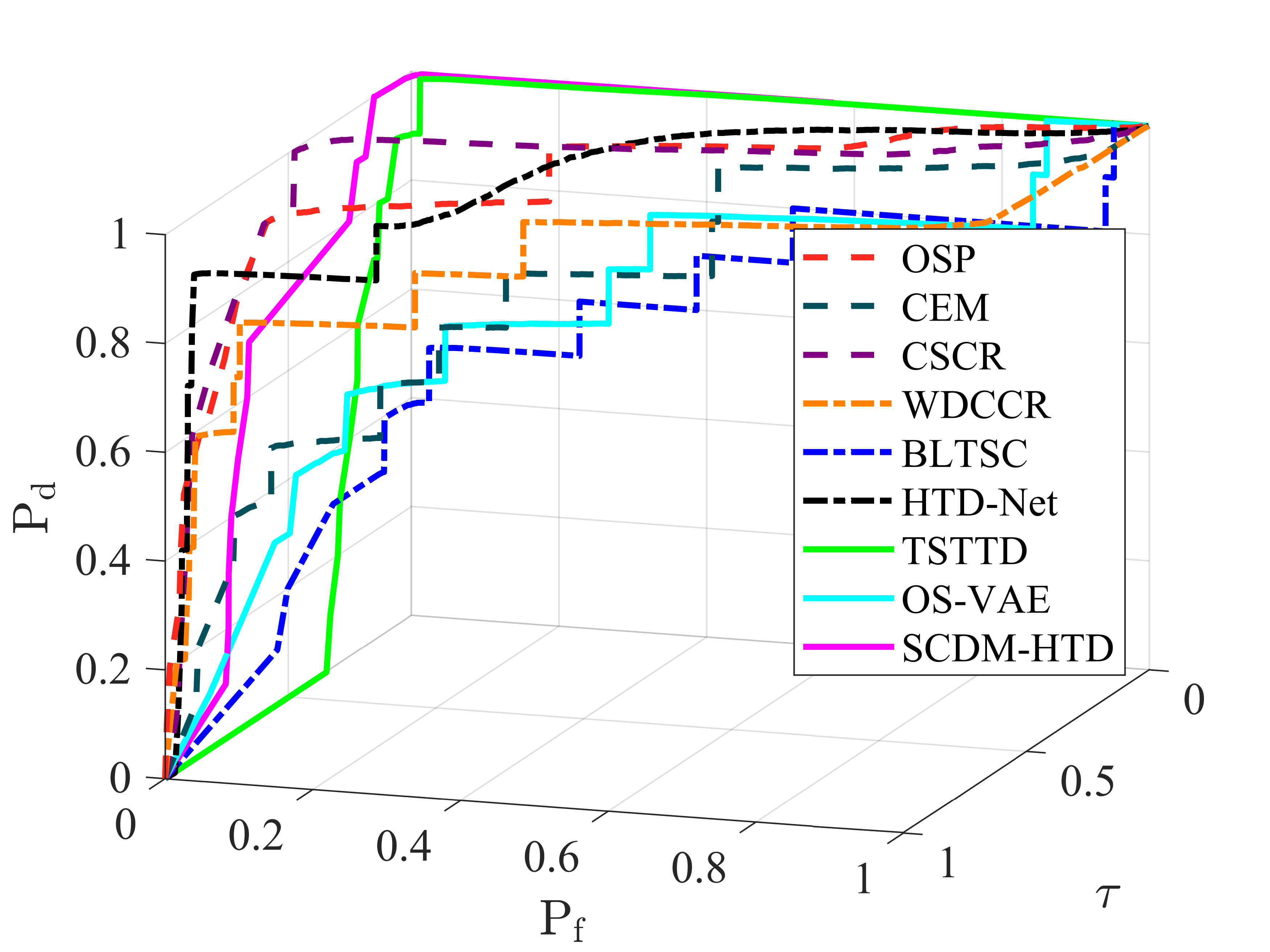}
		}
		\hspace{-18pt}
		\subfigure{
			\includegraphics[width=0.25\textwidth]{figs/tau-pf-pd_Gulfport}
		}
		\hspace{-18pt}
		\subfigure{
			\includegraphics[width=0.25\textwidth]{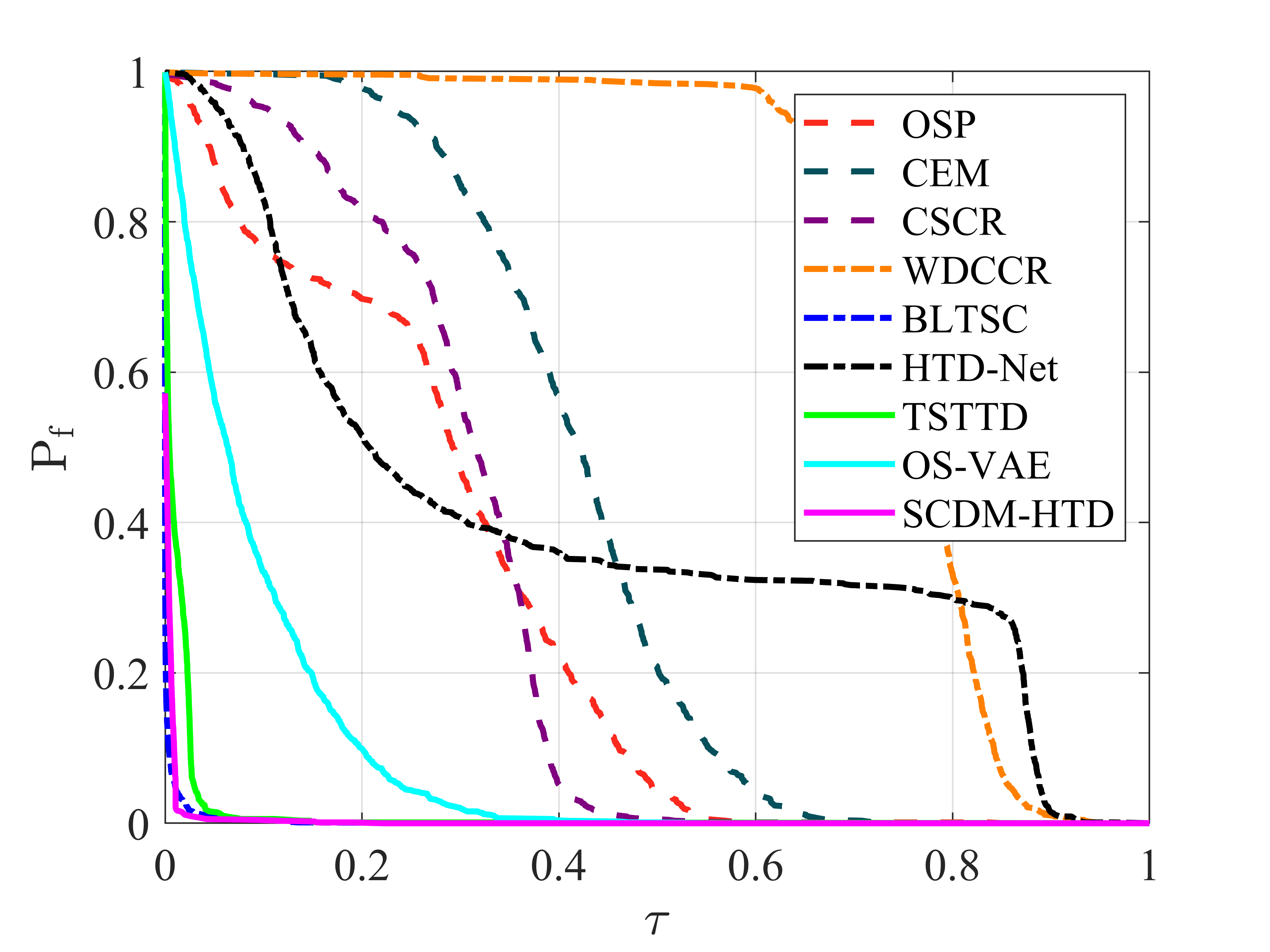}
		}
		\hspace{-18pt}
		\subfigure{
			\includegraphics[width=0.25\textwidth]{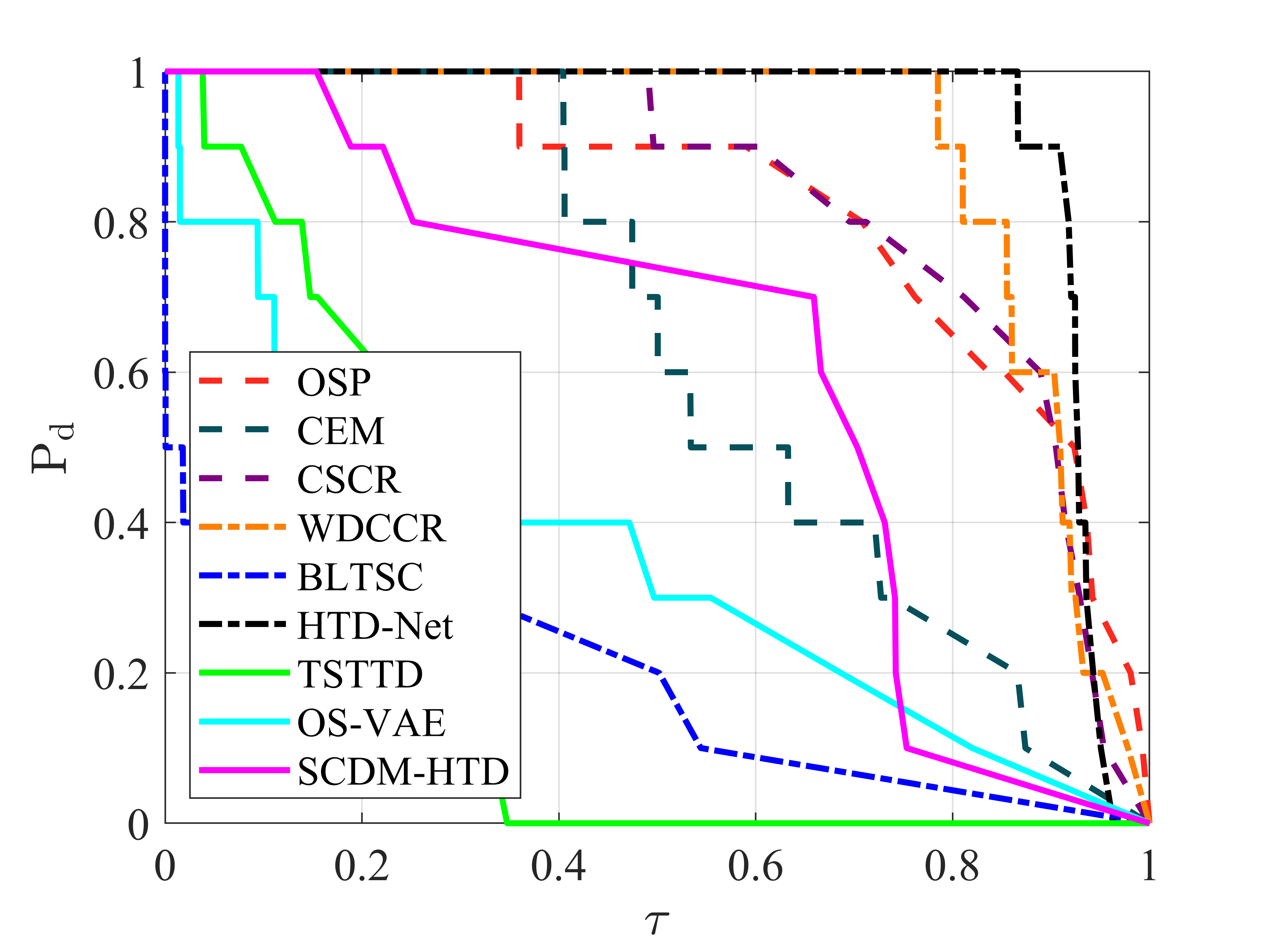}
		}\\
		\parbox{1.0\textwidth}{\centering {\footnotesize (a)}}
	\end{minipage}
	\begin{minipage}{1.0\textwidth}
		\subfigure{
			\includegraphics[width=0.25\textwidth]{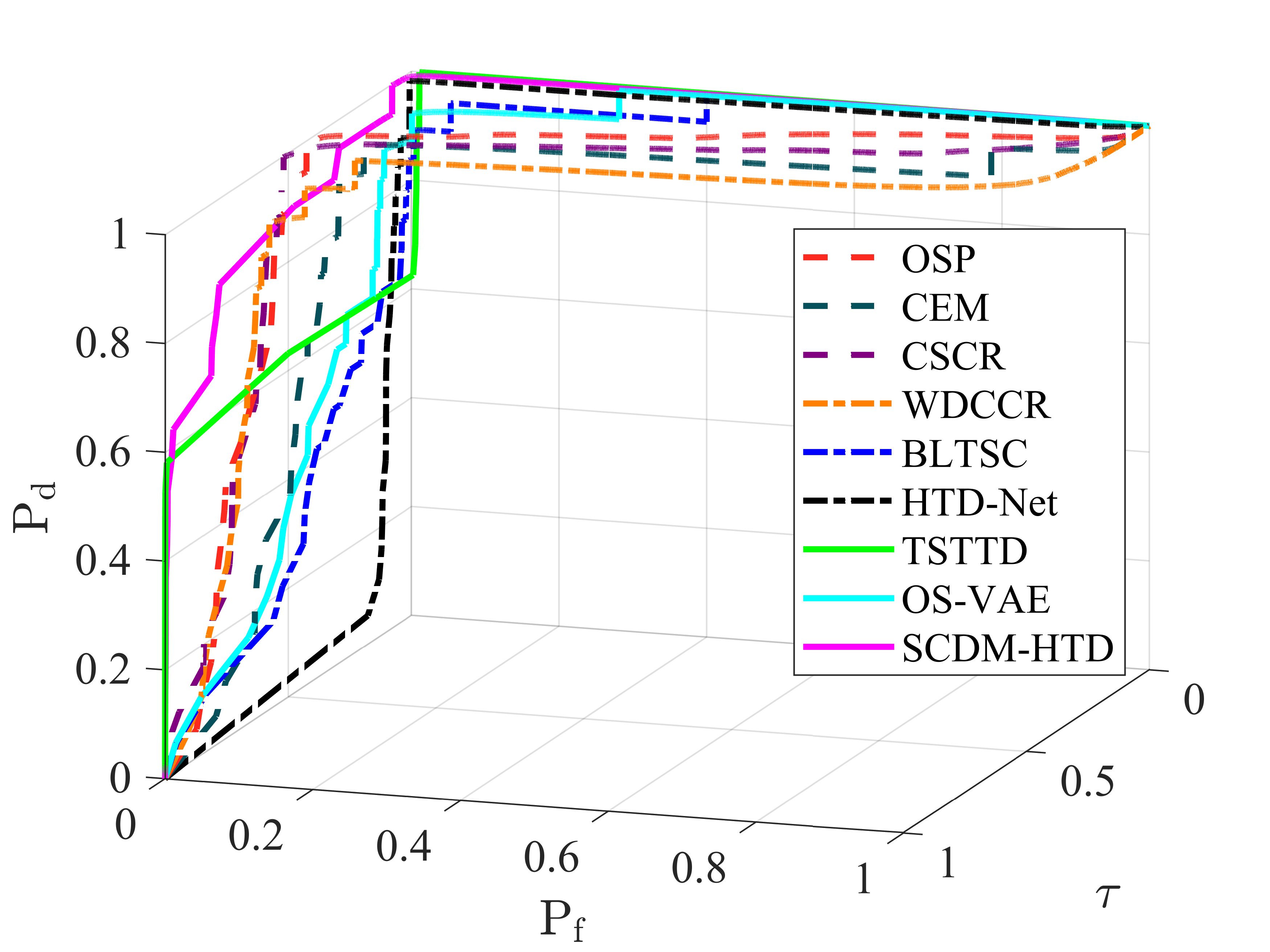}
		}
		\hspace{-18pt}
		\subfigure{
			\includegraphics[width=0.25\textwidth]{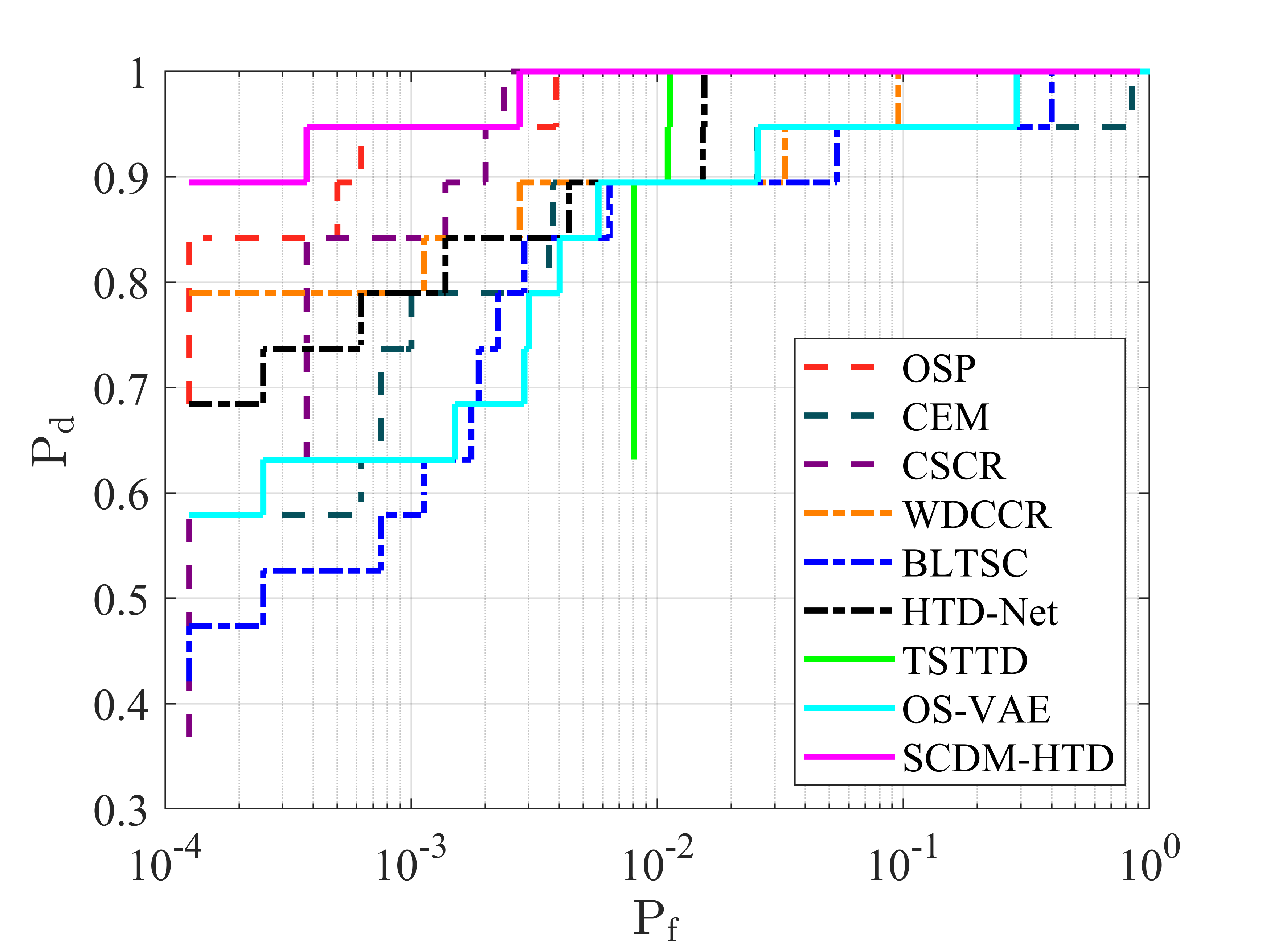}
		}
		\hspace{-18pt}
		\subfigure{
			\includegraphics[width=0.25\textwidth]{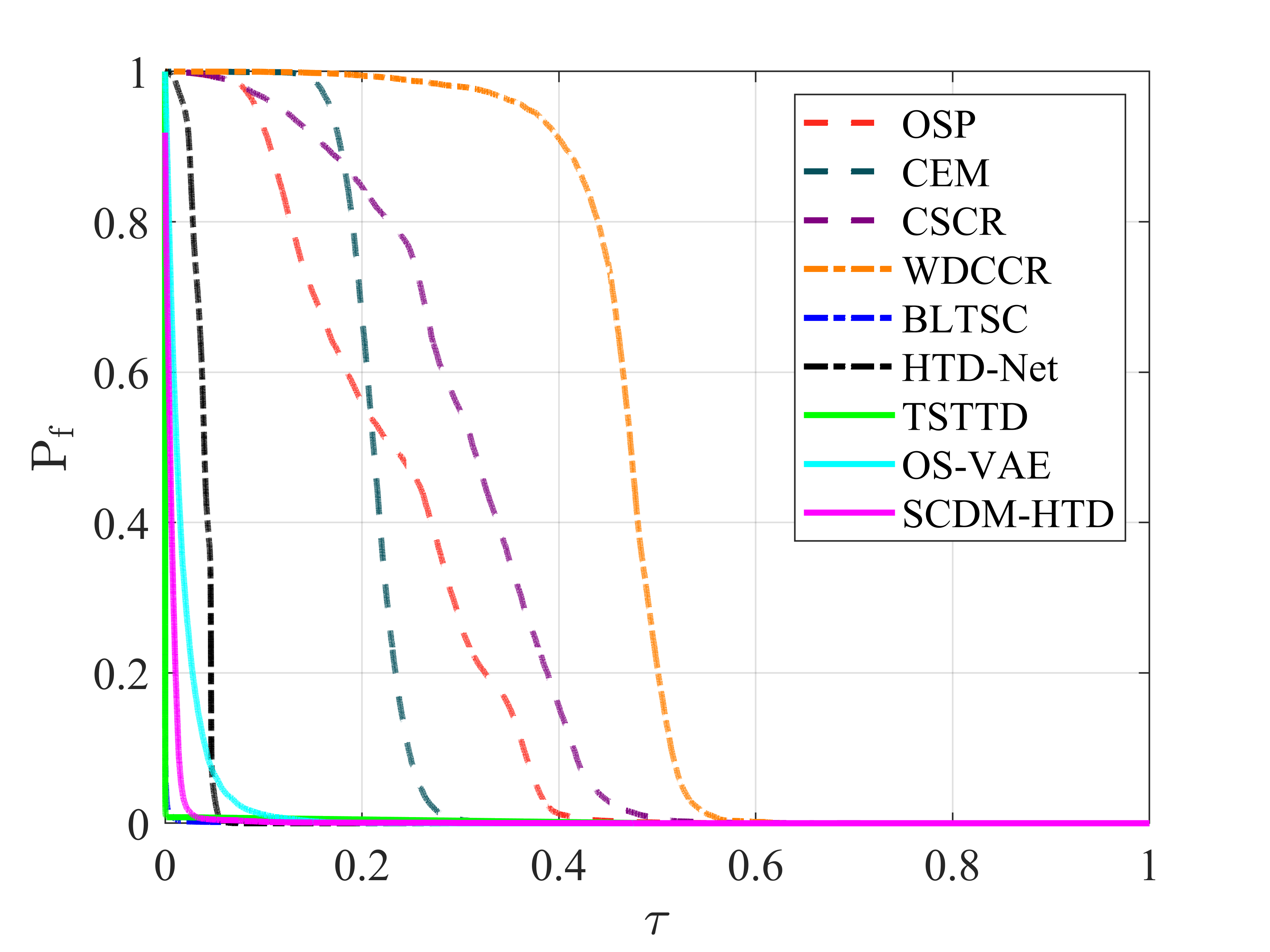}
		}
		\hspace{-18pt}
		\subfigure{
			\includegraphics[width=0.25\textwidth]{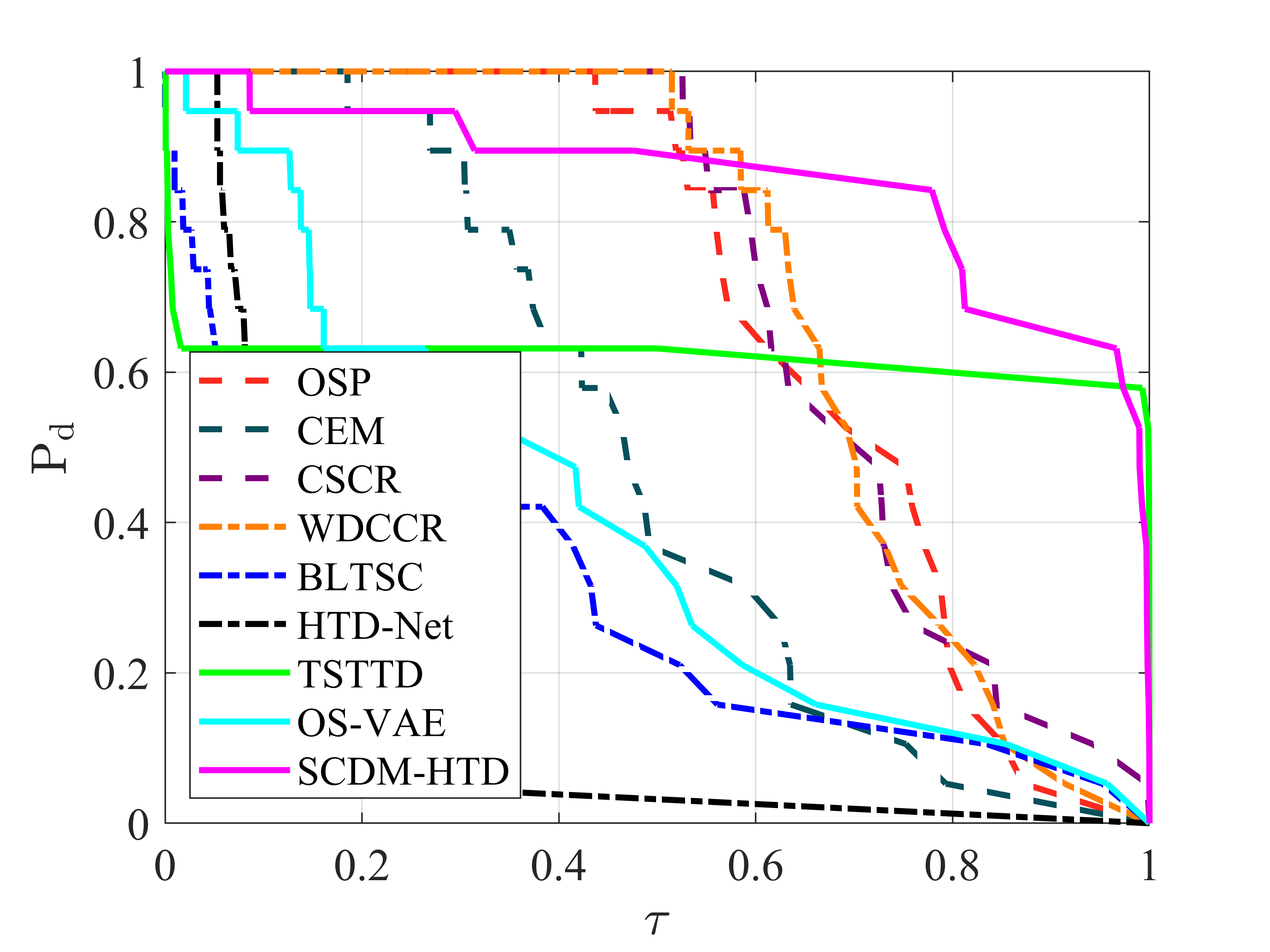}
		}\\
		\parbox{1.0\textwidth}{\centering {\footnotesize (b)}}
	\end{minipage}
	\begin{minipage}{1.0\textwidth}
		\subfigure{
			\includegraphics[width=0.25\textwidth]{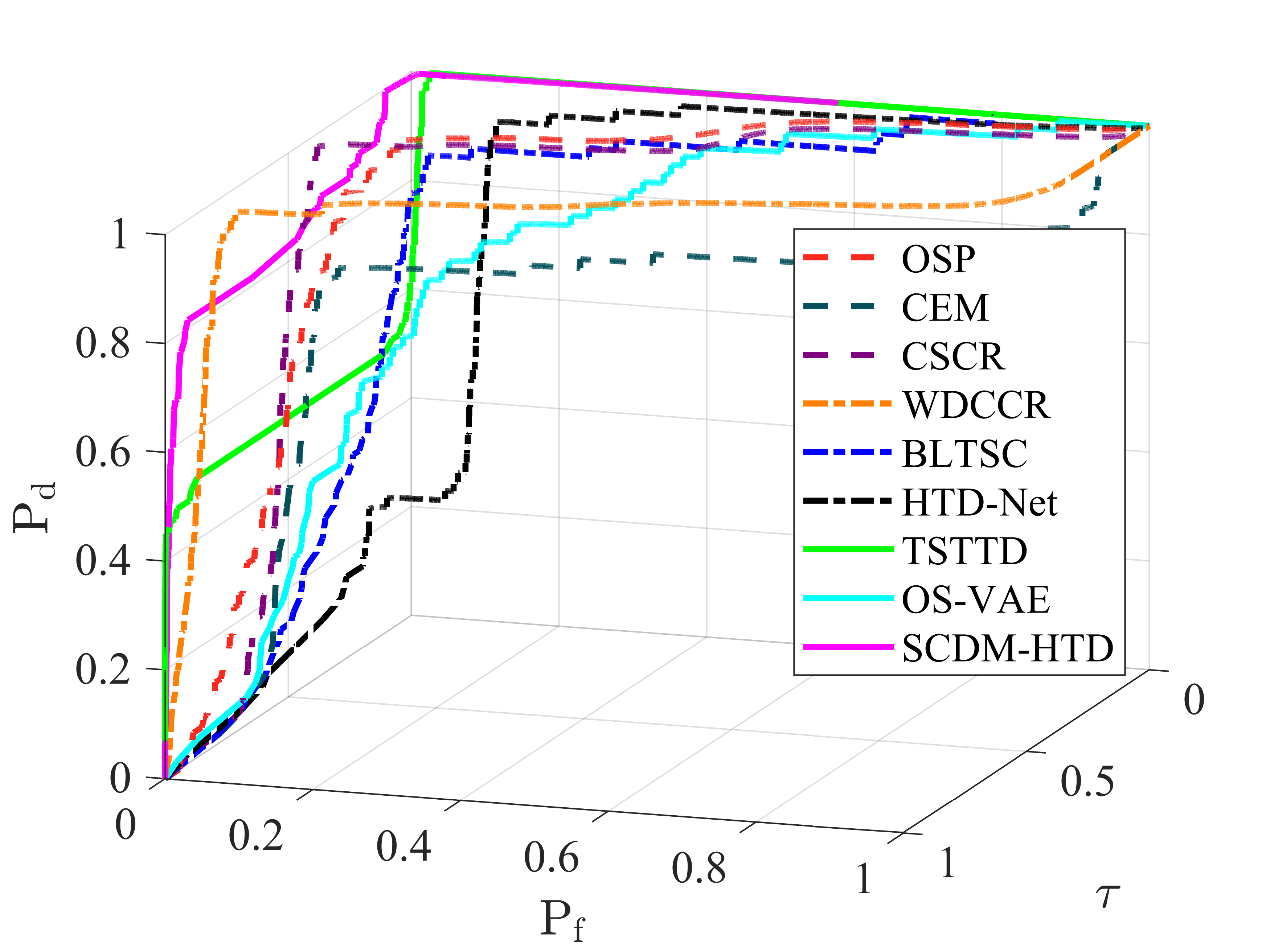}
		}
		\hspace{-18pt}
		\subfigure{
			\includegraphics[width=0.25\textwidth]{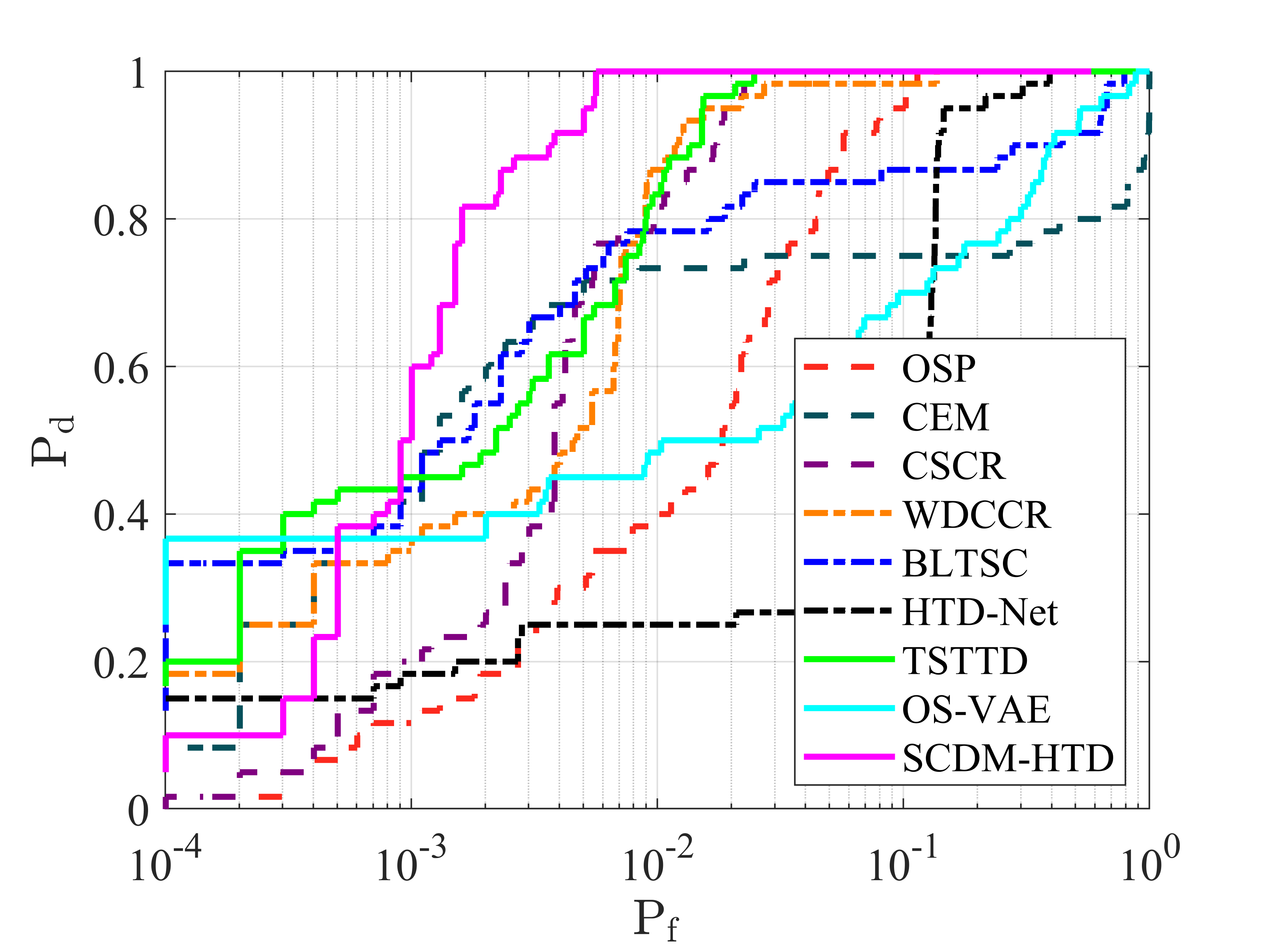}
		}
		\hspace{-18pt}
		\subfigure{
			\includegraphics[width=0.25\textwidth]{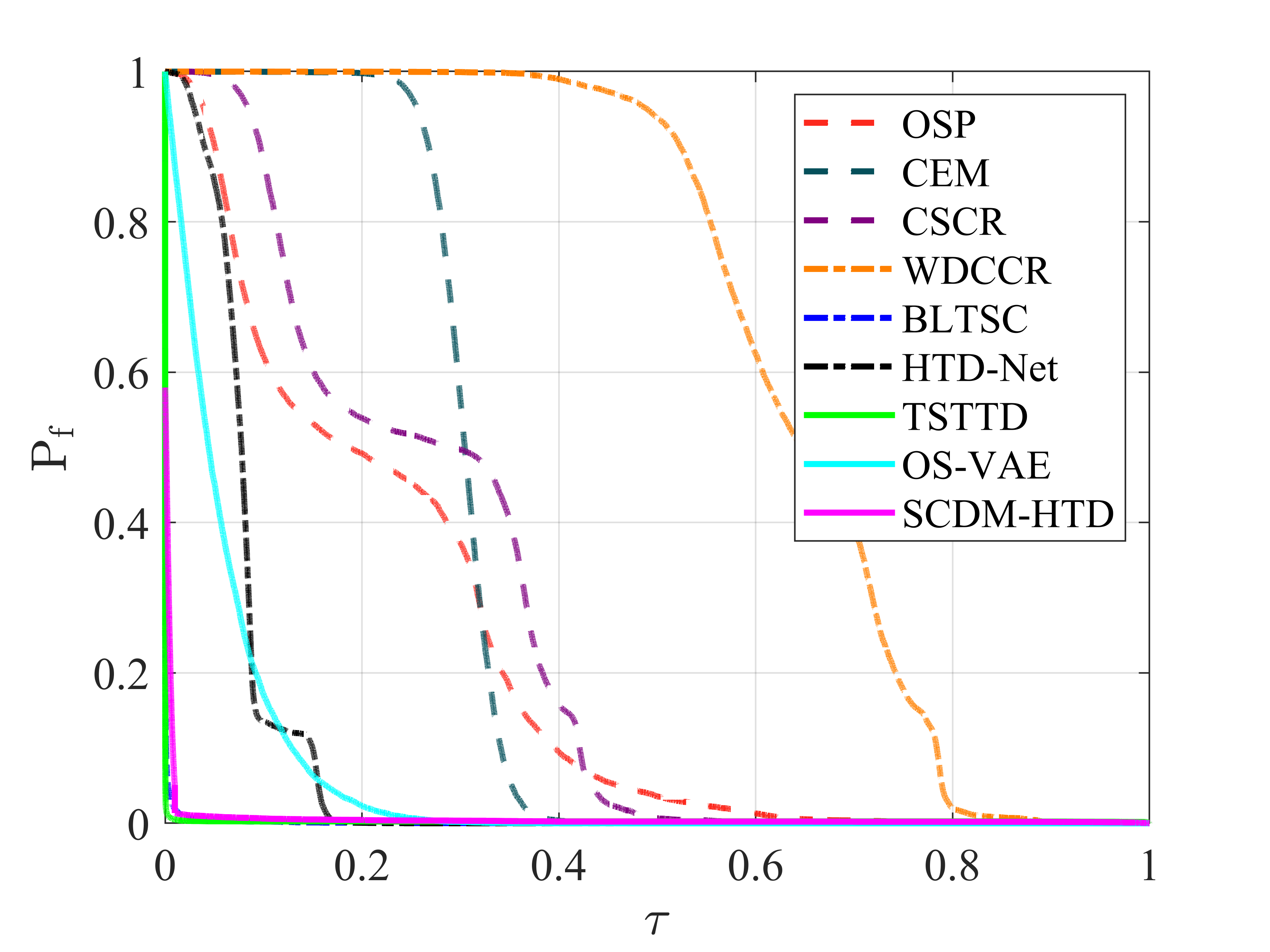}
		}
		\hspace{-18pt}
		\subfigure{
			\includegraphics[width=0.25\textwidth]{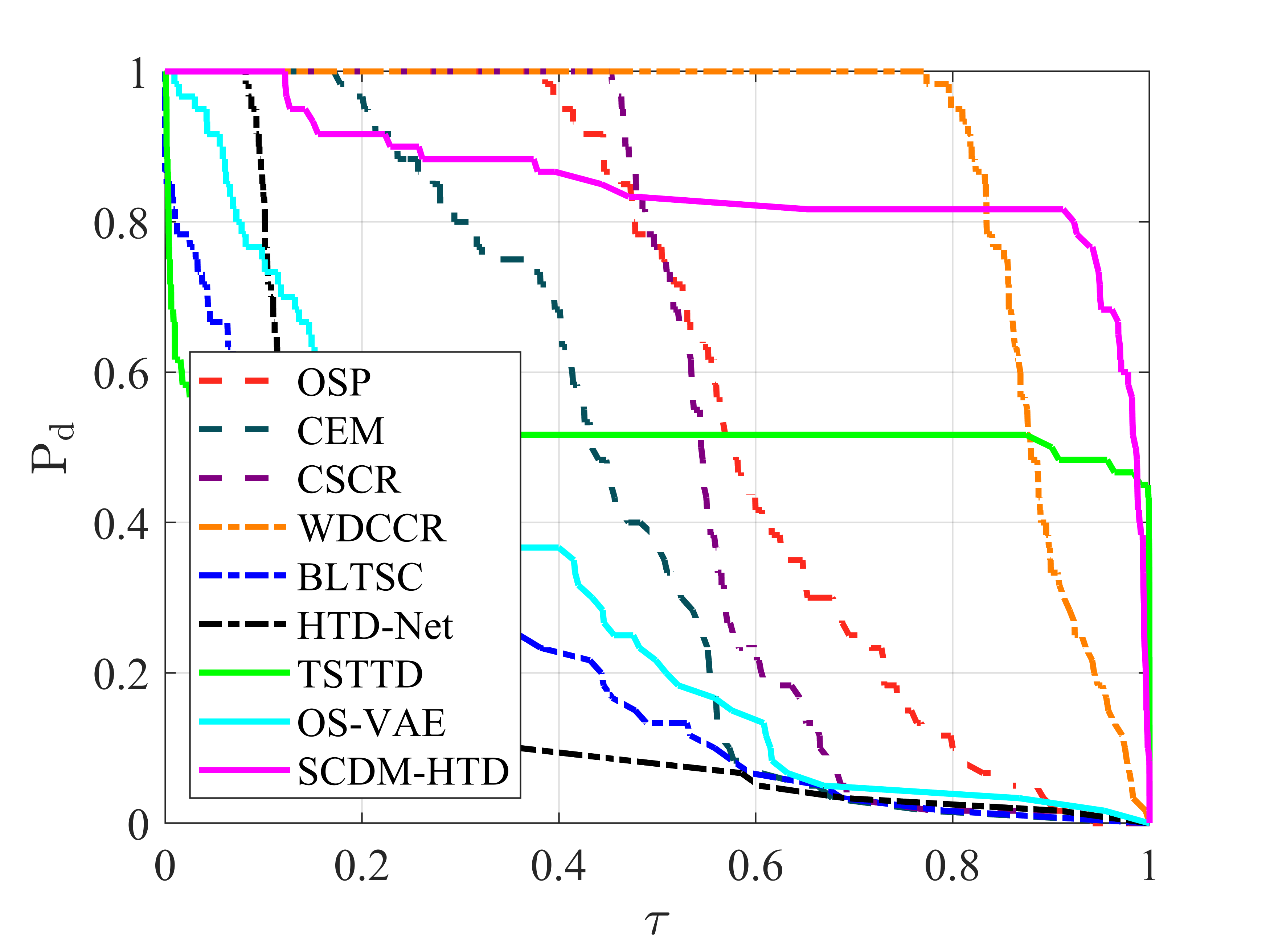}
		}\\
		\parbox{1.0\textwidth}{\centering {\footnotesize (c)}}
	\end{minipage}
	\begin{minipage}{1.0\textwidth}
		\subfigure{
			\includegraphics[width=0.25\textwidth]{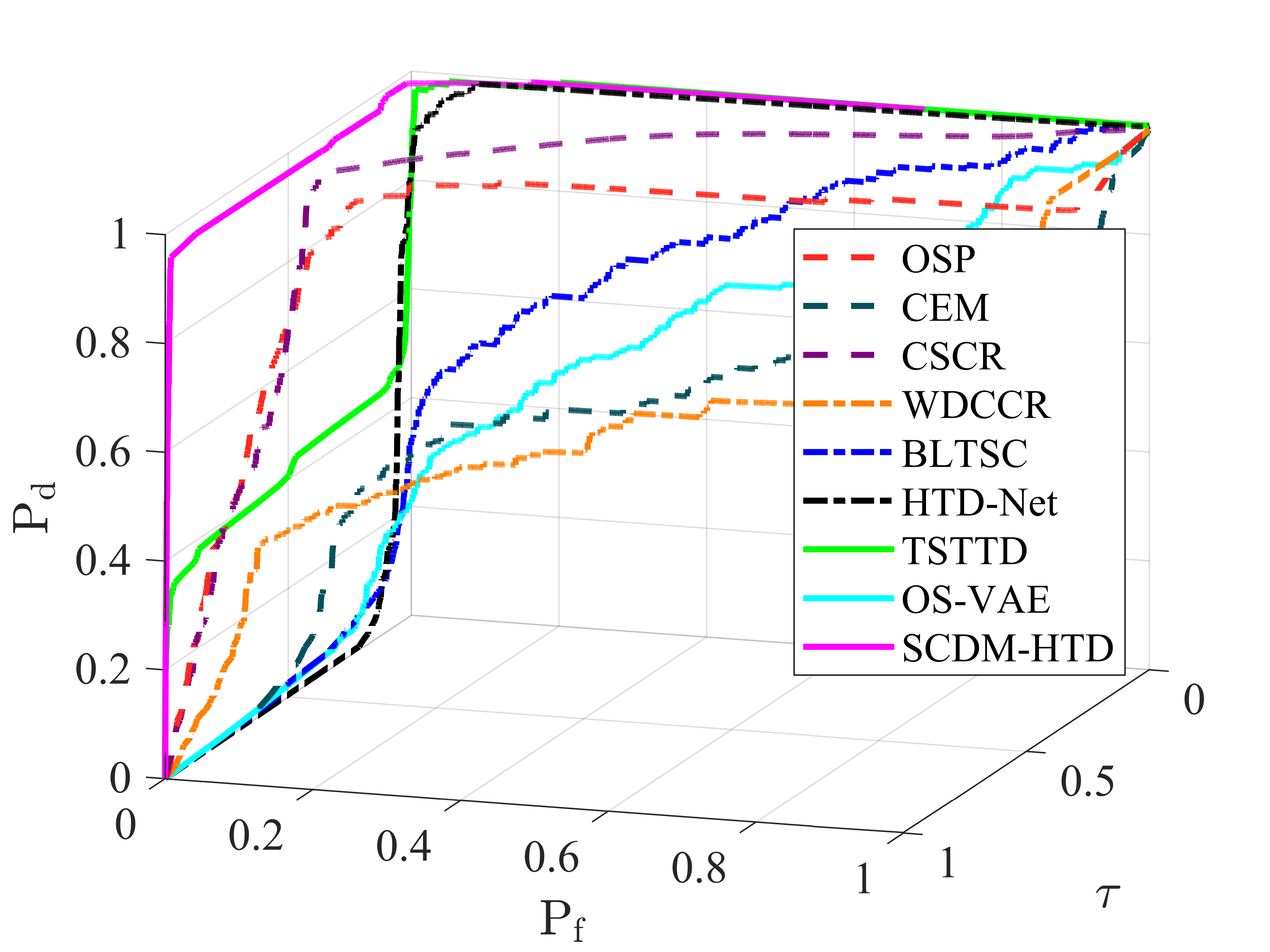}
		}
		\hspace{-18pt}
		\subfigure{
			\includegraphics[width=0.25\textwidth]{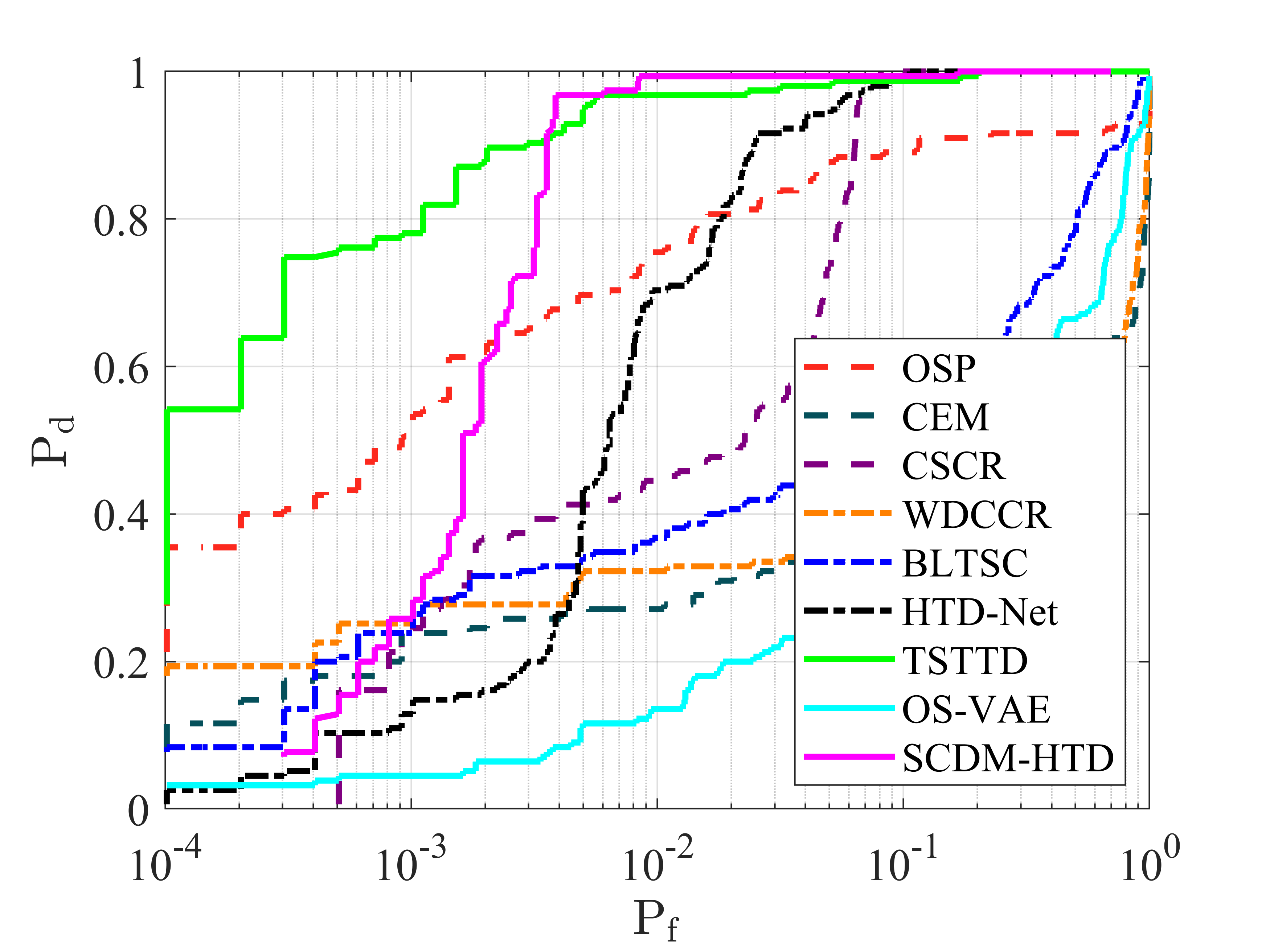}
		}
		\hspace{-18pt}
		\subfigure{
			\includegraphics[width=0.25\textwidth]{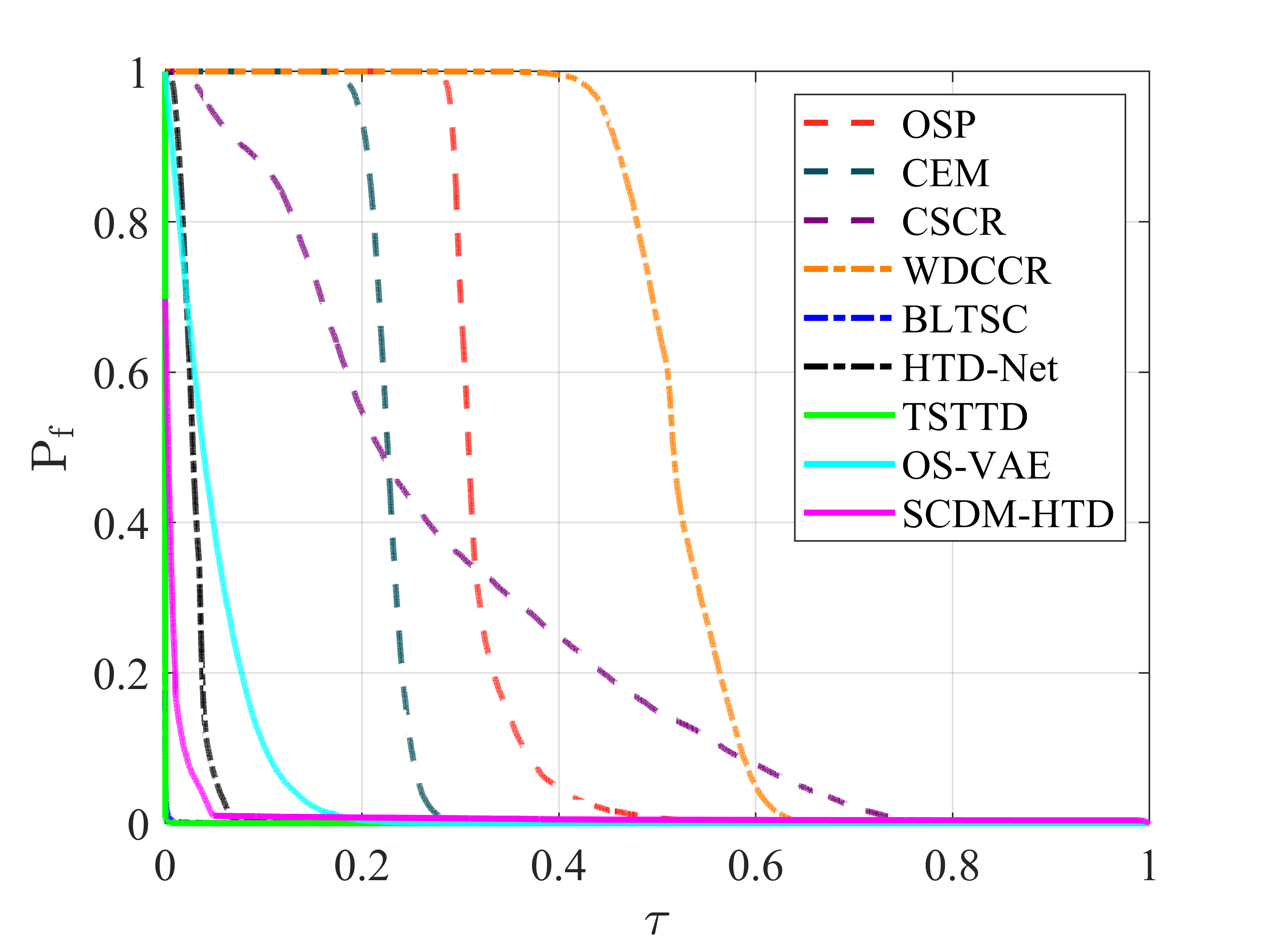}
		}
		\hspace{-18pt}
		\subfigure{
			\includegraphics[width=0.25\textwidth]{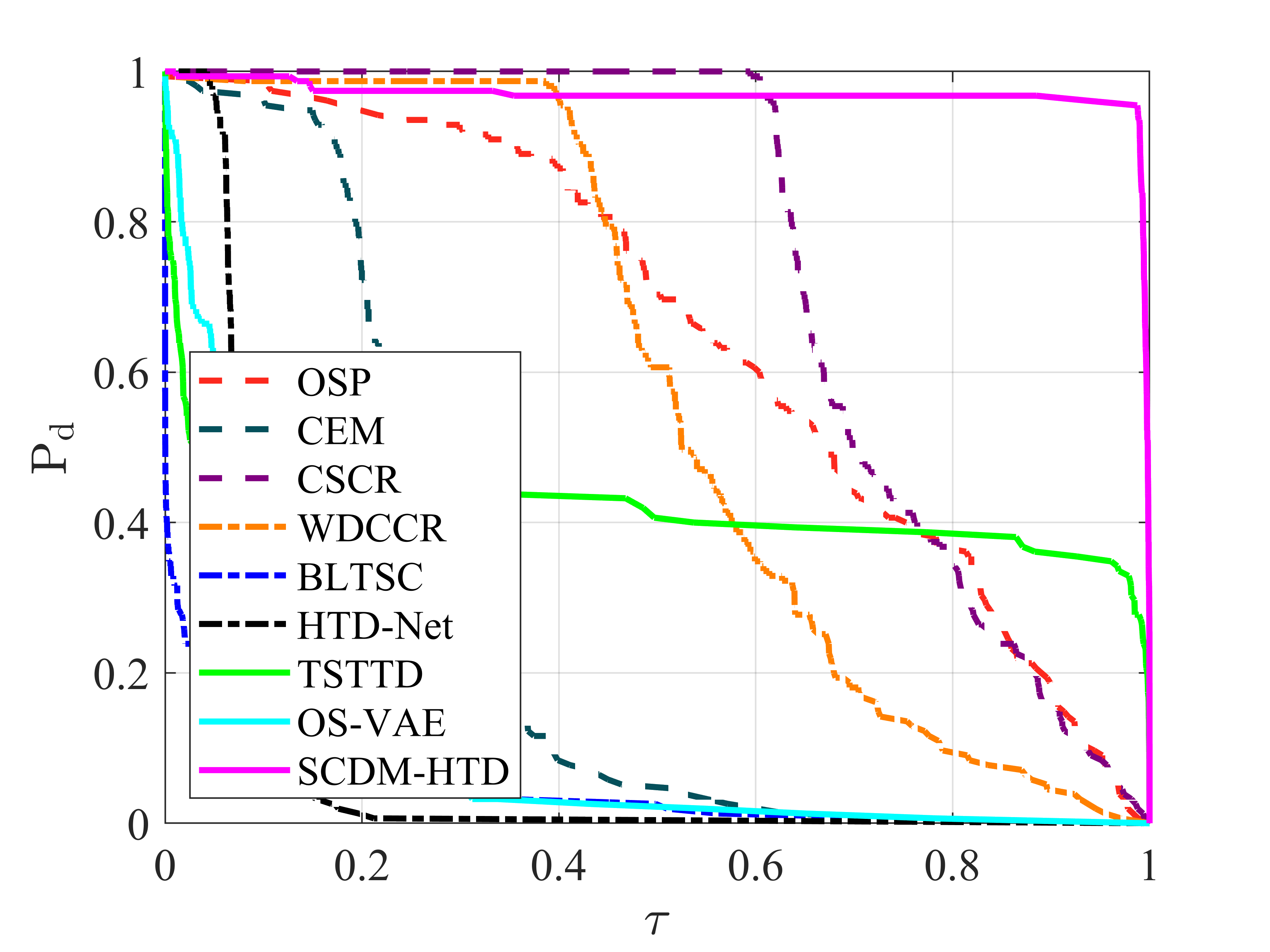}
		}\\
		\parbox{1.0\textwidth}{\centering {\footnotesize (d)}}
	\end{minipage}
	\caption{ROC curves of the competing methods on (a) MULL Gulfport, (b) HYDICE, (c) Airport-4, and (d) Urban-2.  From left to right: 3D ROC curve, 2D ROC curve of $\rm (P_f, P_d)$, 2D ROC curve of $\rm (\tau, P_f)$, and 2D ROC curve of $\rm (\tau, P_d)$.}
	\label{roc_curves_fig}
\end{figure*}
\begin{figure*}[!t]
	\centering
	\subfigure[]{
		\includegraphics[width=0.25\textwidth]{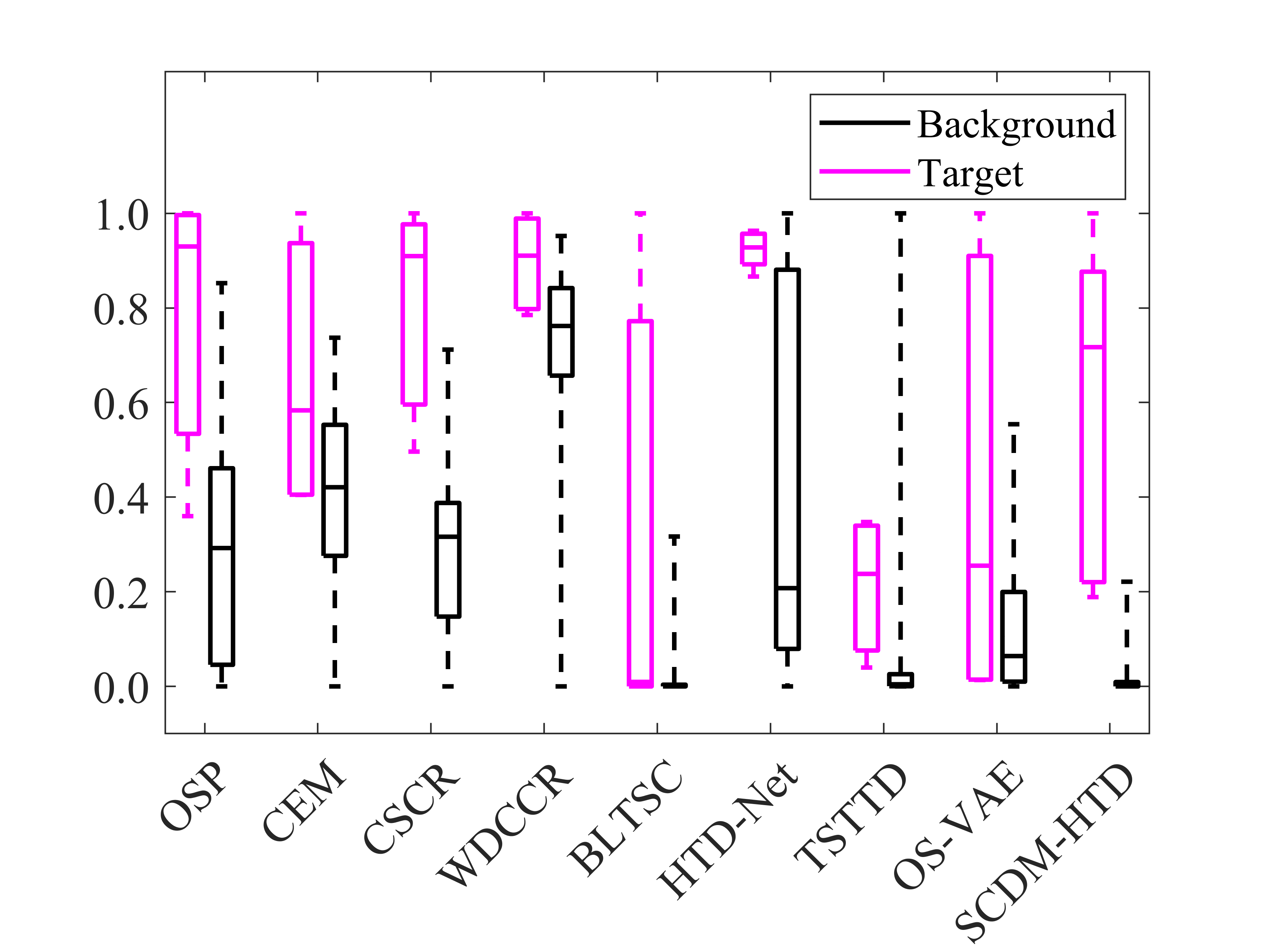}
	}
	\hspace{-18pt}
	\subfigure[]{
		\includegraphics[width=0.25\textwidth]{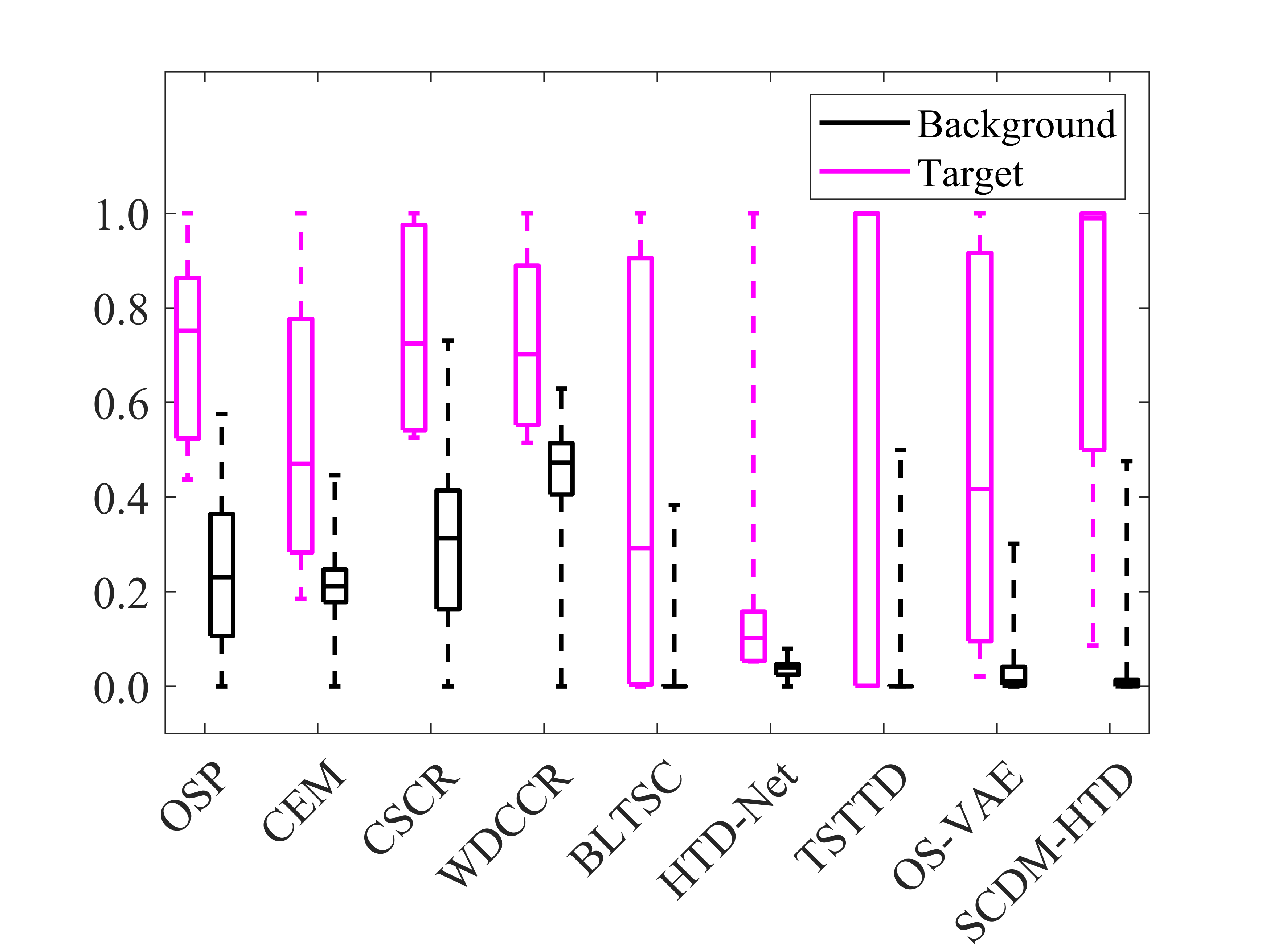}
	}
	\hspace{-18pt}
	\subfigure[]{
		\includegraphics[width=0.25\textwidth]{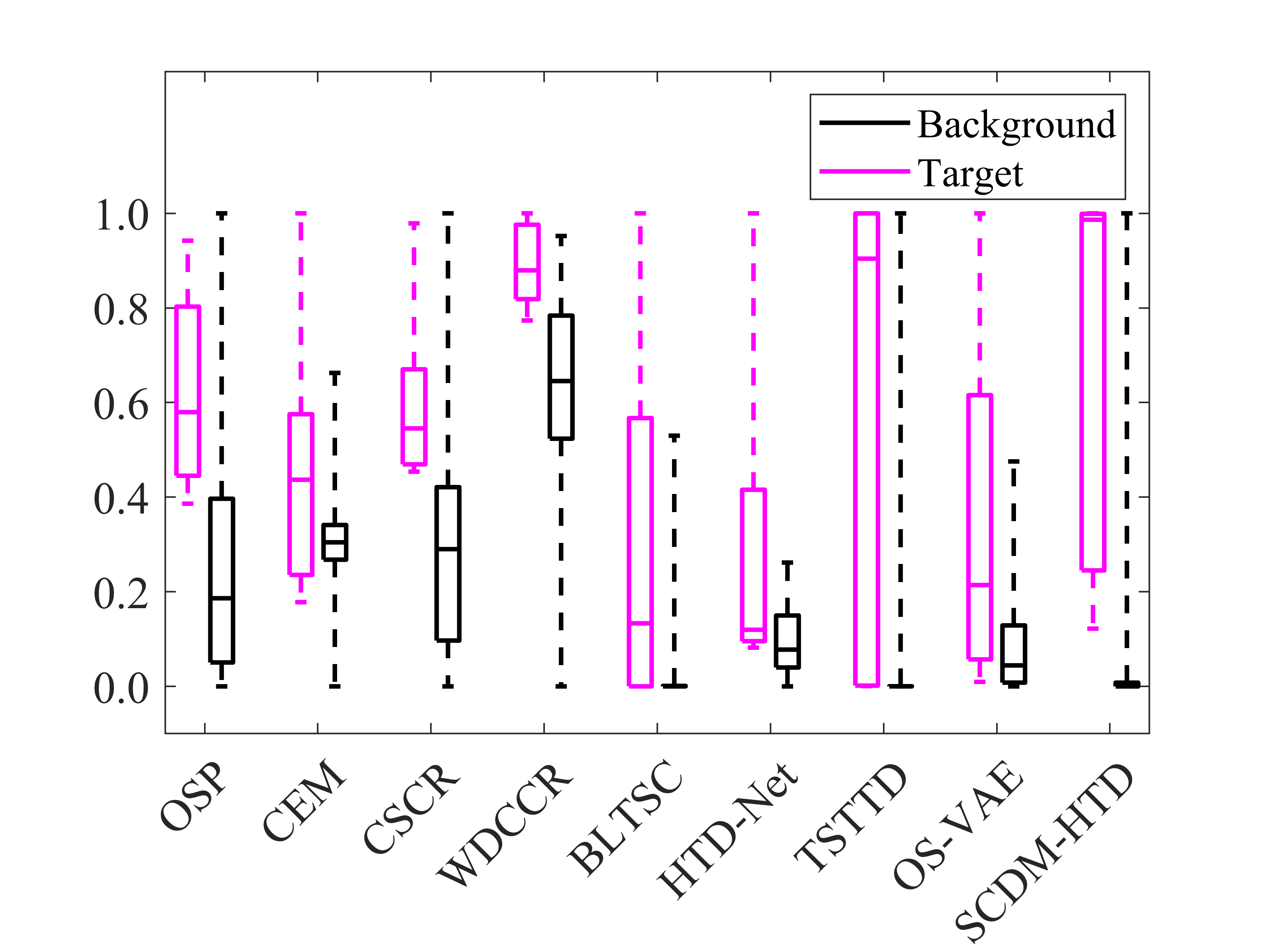}
	}
	\hspace{-18pt}
	\subfigure[]{
		\includegraphics[width=0.25\textwidth]{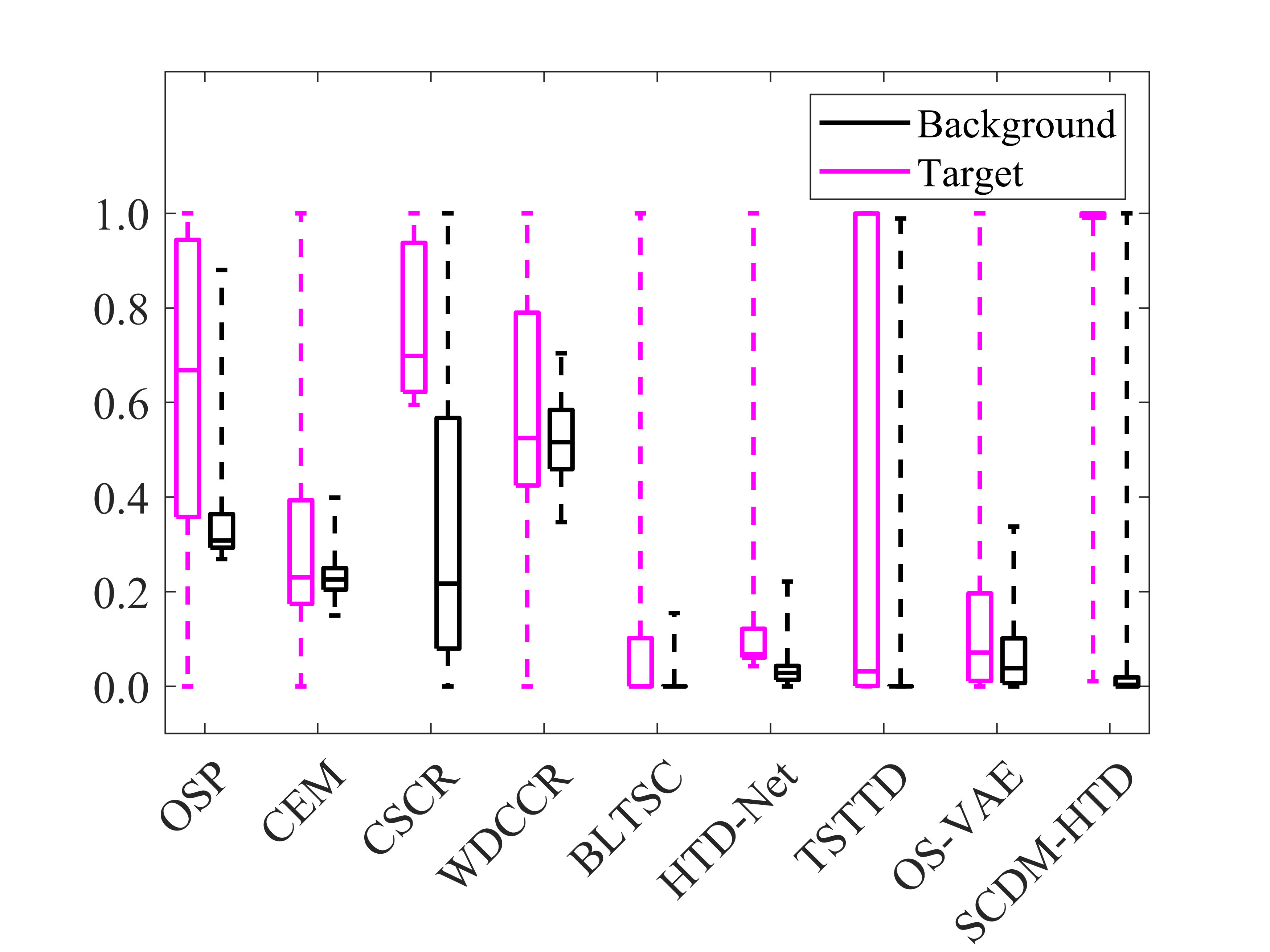}
	}
	\caption{Background-target separability diagrams of competing methods on four datasets. (a) MULL Gulfport. (b) HYDICE. (c) Airport-4. (d) Urban-2.}
	\label{background_target_separation}
\end{figure*}
To evaluate the qualitative performance of SCDM-HTD, we visualize the detection maps in Fig. \ref{detection_results}. All competing methods achieve some level of detection performance. However, OSP, CEM, CSCR, and WDCCR exhibit weak background suppression, while BLTSC, HTD-Net, TSTTD, and OS-VAE tend to miss the targets to varying degrees. In contrast, the proposed SCDM-HTD maintains a promising balance between target saliency and background suppression. Fig. \ref{roc_curves_fig} presents the ROC curves for the competing methods. The 2D ROC curves of $\rm (P_f, P_d)$, $\rm (\tau, P_f)$, and $\rm (\tau, P_d)$ for SCDM-HTD are predominantly located in the upper-left, lower-left, and upper-right corners, respectively. This demonstrates its strong performance in detection effectiveness, background suppression, and target detectability. Fig. \ref{background_target_separation} displays the background-target separability diagrams of the competing methods. Compared to other methods, SCDM-HTD exhibits a larger margin between the background and target boxes, highlighting its outstanding discrimination ability. Additionally, the smaller background box further emphasizes its superior background suppression. 
\subsubsection{Quantitative Comparison} To assess the quantitative performance of the proposed SCDM-HTD, Table \ref{auc_results} presents the AUC results for the competing methods. SCDM-HTD achieves the highest score on the key metric, $\rm{AUC}_{(P_f, P_d)}$, confirming its superior detection effectiveness. It also demonstrates competitive performance across $\rm{AUC}_{(\tau, P_f)}$, $\rm{AUC}_{(\tau, P_d)}$, $\rm{AUC}_{\rm{OA}}$, and $\rm{AUC}_{\rm{SNPR}}$, highlighting its consistent reliability for high-confidence detection. Table \ref{auc_results} also reports the running times of all competing methods. Statistical-based methods such as OSP and CEM are the fastest due to their simplified computations. The proposed SCDM-HTD, while maintaining high performance, also achieves low computational cost thanks to its lightweight model and the accelerated reverse sampling process. This efficiency makes it particularly well-suited for practical applications.
\begin{table*}[!t]
	\centering
	\caption{AUC SCORES AND RUNNING TIME (IN SECONDS) OF COMPETING METHODS ON THE FOUR DATASETS. BOLD REPRESENTS THE BEST RESULT WHILE UNDERLINED THE SECOND.}
	\scriptsize
	\tabcolsep = 9pt
	\renewcommand\arraystretch{1.5}
	\begin{tabular}{clccccccccc}
		\hline
		Dataset&&OSP&CEM&CSCR&WDCCR&BLTSC&HTD-Net&TSTTD&OS-VAE&SCDM-HTD\\ 
		%\midrule  % 中部线
		%\vspace{0.5em}
		\hline
		\multirow{6}*{MULL Gulfport}&$\rm{AUC}_{(P_f, P_d)}$&0.9690&0.8257&\underline{0.9993}&0.9184&0.6943&0.9719&0.9958&0.7501&\bf{0.9999}\\
		&$\rm{AUC}_{(\tau, P_d)}$&0.8553&0.6493&0.8646&\underline{0.8981}&0.2435&\bf{0.9314}&0.2666&0.3934&0.6828\\
		&$\rm{AUC}_{(\tau, P_f)}$&0.2724&0.4162&0.2960&0.7510&\bf{0.0028}&0.3960&0.0122&0.0873&\underline{0.0039}\\
		&$\rm{AUC}_{\rm{OA}}$&1.5519&1.0588&\underline{1.5679}&1.0655&0.9350&1.5073&1.2502&1.0562&\bf{1.6788}\\
		&$\rm{AUC}_{\rm{SNPR}}$&3.1399&1.5601&2.9209&1.1959&\underline{86.9643}&2.3520&21.8525&4.5063&\bf{175.0769}\\
		&Running time&\underline{0.0090}&\bf{0.0061}&0.1357&2.2048&1.7499&4.0861&1.7609&0.4708&0.3357\\
		\hline
		\multirow{6}*{HYDICE}&$\rm{AUC}_{(P_f, P_d)}$&\underline{0.9997}&0.9535&0.9996&0.9930&0.9752&0.9980&0.9967&0.9825&\bf{0.9998}\\
		&$\rm{AUC}_{(\tau, P_d)}$&0.7128&0.5186&\underline{0.7281}&0.7278&0.3493&0.1665&0.6462&0.4316&\bf{0.8746}\\
		&$\rm{AUC}_{(\tau, P_f)}$&0.2293&0.2125&0.3024&0.4643&\bf{0.0006}&0.0379&\underline{0.0060}&0.0183&0.0068\\
		&$\rm{AUC}_{\rm{OA}}$&1.4832&1.2596&1.4253&1.2565&1.3239&1.1266&\underline{1.6369}&1.3958&\bf{1.8676}\\
		&$\rm{AUC}_{\rm{SNPR}}$&3.1086&2.4405&2.4077&1.5675&\bf{582.1666}&4.3931&107.7000&23.5847&\underline{128.6176}\\
		&Running time&\bf{0.0240}&\underline{0.0499}&2.4220&20.6579&1.8143&23.0791&2.6125&1.1417&0.3552\\
		\hline
		\multirow{6}*{Airport-4}&$\rm{AUC}_{(P_f, P_d)}$&0.9753&0.7971&0.9942&0.9927&0.9234&0.8972&\underline{0.9954}&0.8714&\bf{0.9986}\\
		&$\rm{AUC}_{(\tau, P_d)}$&0.6063&0.4481&0.5630&\bf{0.8895}&0.2289&0.1968&0.5254&0.3042&\underline{0.8554}\\
		&$\rm{AUC}_{(\tau, P_f)}$&0.2137&0.3045&0.2559&0.6472&\bf{0.0012}&0.0798&\underline{0.0026}&0.0586&0.0058\\
		&$\rm{AUC}_{\rm{OA}}$&1.3679&0.9407&1.3013&1.2350&1.1511&1.0142&\underline{1.5182}&1.1170&\bf{1.8482}\\
		&$\rm{AUC}_{\rm{SNPR}}$&2.8372&1.4716&2.2001&1.3744&\underline{190.7500}&2.4662&\bf{202.0769}&5.1911&147.4828\\
		&Running time&\bf{0.0439}&\underline{0.0635}&6.4061&26.8052&2.4552&30.6488&3.0015&1.0733&0.3845\\
		\hline
		\multirow{6}*{Urban-2}&$\rm{AUC}_{(P_f, P_d)}$&0.9113&0.5508&0.9730&0.5599&0.7647&0.9875&\underline{0.9963}&0.6434&\bf{0.9969}\\
		&$\rm{AUC}_{(\tau, P_d)}$&0.6561&0.2661&\underline{0.7482}&0.5743&0.0500&0.0829&0.4232&0.1030&\bf{0.9698}\\
		&$\rm{AUC}_{(\tau, P_f)}$&0.3204&0.2269&0.2778&0.5201&\underline{0.0002}&0.0292&\underline{0.0002}&0.0484&0.0122\\
		&$\rm{AUC}_{\rm{OA}}$&1.2470&0.5900&\underline{1.4434}&0.6141&0.8145&1.0412&1.4193&0.6980&\bf{1.9545}\\
		&$\rm{AUC}_{\rm{SNPR}}$&2.0478&1.1728&2.6933&1.1042&\underline{250.0000}&2.8390&\bf{2116.0000}&2.1281&79.4918\\
		&Running time&\bf{0.0675}&\underline{0.0698}&1.6889&32.7765&2.6663&32.32560&3.1734&1.0574&0.3826\\
		\hline
	\end{tabular}
	\label{auc_results}
\end{table*}
\subsection{Parameter Analysis}
\begin{figure}[!t]
	\centering
	\subfigure[]{
		\includegraphics[width=0.22\textwidth]{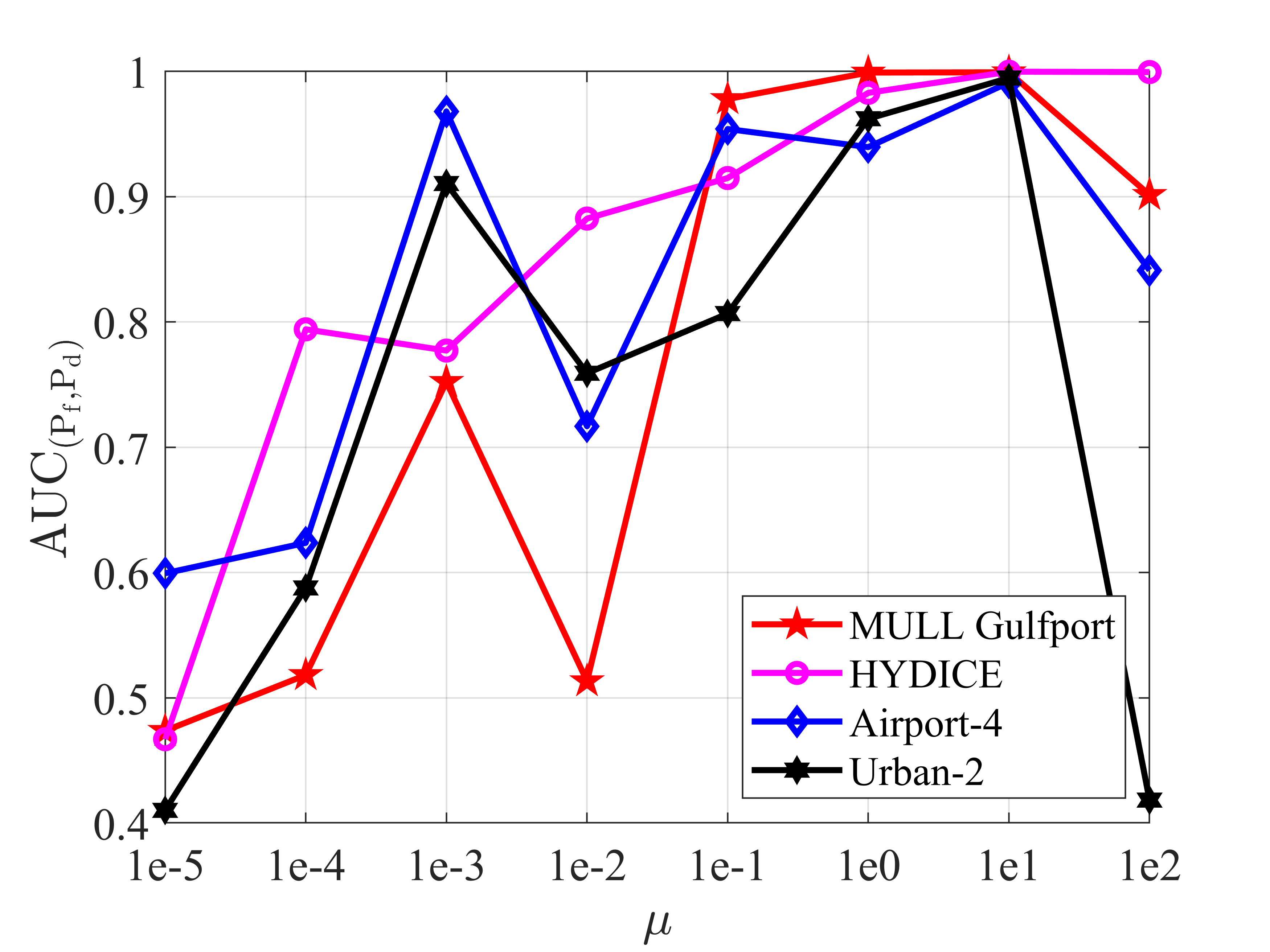}
	}
	\subfigure[]{
		\includegraphics[width=0.22\textwidth]{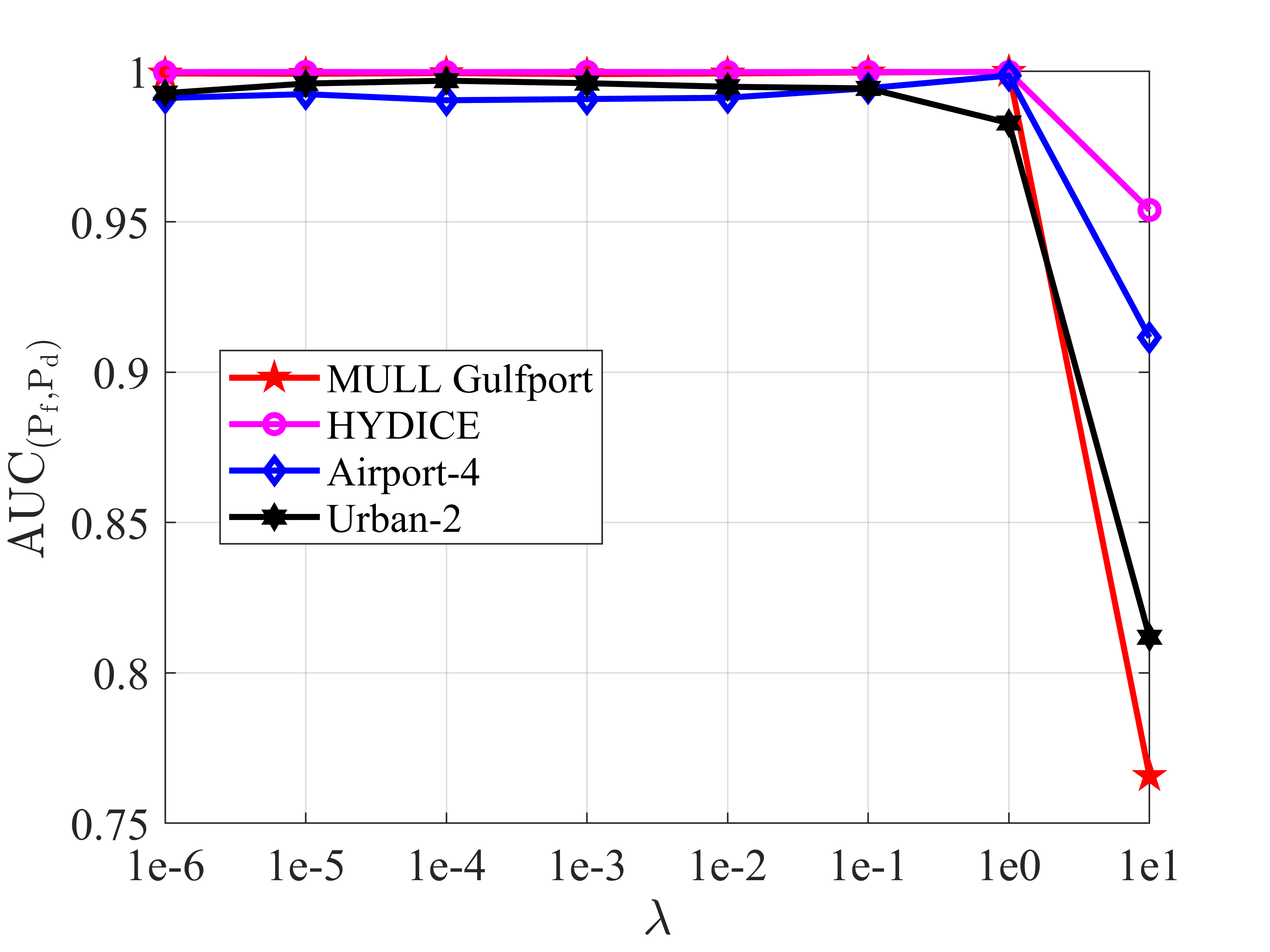}
	}
	\caption{Effect of parameters. (a) $\mu$. (b) $\lambda$. }
	\label{parameter_analysis}
\end{figure}
\subsubsection{Effect of Parameter $\mu$} The parameter $\mu$ balances background suppression and target enhancement in the CEM loss. To analyze its impact, we plot $\rm{AUC}_{(P_f, P_d)}$ as a function of $\mu$ over the range $10^{-5}$ to $100$. As shown in Fig. \ref{parameter_analysis}-(a), detection performance generally improves initially, but starts to decline as $\mu$ increases across the four datasets. The optimal performance is consistently achieved at $\mu = 10$, so we set $\mu$ to 10 for all datasets.
\subsubsection{Effect of Parameter $\lambda$} The parameter $\lambda$ controls the weight of the BCE loss. To evaluate its impact, we plot $\rm{AUC}_{(P_f, P_d)}$ as a function of $\lambda$ over the range $10^{-6}$ to $10$. As shown in Fig. \ref{parameter_analysis}-(b), detection performance remains stable or improves slightly at first, but begins to decline as $\lambda$ increases further. Based on the peak values, we set $\lambda$ to 1, $10^{-3}$, 1, and $10^{-4}$ for MULL Gulfport, HYDICE, Airport-4, and Urban-2, respectively.
\subsection{Ablation Study}
In this section, we evaluate the contributions of the background-target separation loss and the reverse sampling acceleration to the overall performance of the proposed method.
\begin{table}[!t]
	\centering
	\caption{Effectiveness of background-target separation loss on four datasets. Bold represents the best result.}
	\scriptsize
	\tabcolsep = 6pt
	\renewcommand\arraystretch{1.5}
	\begin{tabular}{ccccc}
		\hline
		Loss&MULL Gulfport&HYDICE&Airport-4&Urban-2\\ 
		\hline
		%\midrule  % 中部线
		%\vspace{0.5em}
		Diffusion loss&0.7804&0.9947&0.9669&0.9499\\
		Separation loss&{\bf 0.9999}&{\bf 0.9998}&{\bf 0.9986}&{\bf 0.9969}\\
		\hline
	\end{tabular}
	\label{effect_sep_loss}
\end{table}
\begin{table}[!t]
	\centering
	\caption{Effectiveness of CEM loss and BCE loss on four datasets. Bold represents the best result.}
	\scriptsize
	\tabcolsep = 9pt
	\renewcommand\arraystretch{1.5}
	\begin{tabular}{cccc}
		\hline
		Datasets&CEM loss&BCE loss&${\rm AUC_{(P_f, P_d)}}$\\ 
		%\midrule  % 中部线
		%\vspace{0.5em}
		\hline
		\multirow{3}*{MULL Gulfport}
		&&\checkmark&0.9996\\
		&\checkmark&&0.9993\\
		&\checkmark&\checkmark&{\bf 0.9999}\\
		\hline
		\multirow{3}*{HYDICE}
		&&\checkmark&0.8646\\
		&\checkmark&&0.9996\\
		&\checkmark&\checkmark&{\bf 0.9998}\\
		\hline
		\multirow{3}*{Airport-4}
		&&\checkmark&0.8843\\
		&\checkmark&&0.9912\\
		&\checkmark&\checkmark&{\bf 0.9986}\\
		\hline
		\multirow{3}*{Urban-2}
		&&\checkmark&0.9673\\
		&\checkmark&&0.9948\\
		&\checkmark&\checkmark&{\bf 0.9969}\\
		\hline
	\end{tabular}
	\label{ablation_cem_bce}
\end{table}
\subsubsection{Effectiveness of Background-Target Separation Loss}
The background-target separation loss combines the CEM and BCE losses to improve the distinction between background and target. To assess its effectiveness, we first compare it with the diffusion loss. As shown in Table \ref{effect_sep_loss}, the proposed background-target separation loss significantly outperforms the diffusion loss. This improvement arises because the diffusion loss only reconstructs confidence scores based on cosine similarity, whereas the separation loss not only emphasizes the target but also suppresses the background, effectively differentiating them into distinct classes.

Further, we analyze the individual contributions of CEM and BCE losses through ablation experiments. As shown in Table \ref{ablation_cem_bce}, both CEM and BCE losses achieve reasonable detection performance independently, but their combination results in optimal performance. This is due to the complementary roles of the losses: CEM loss enhances target discrimination and background suppression, while BCE loss uses pixel-wise labels to separate confidence scores into background and target classes. Together, the separation loss significantly improves detection robustness.
\begin{table*}[!t]
	\centering
	\caption{Effectiveness of reverse sampling acceleration on four datasets. Bold represents the best result.}
	\scriptsize
	\tabcolsep = 10pt
	\renewcommand\arraystretch{1.5}
	\begin{tabular}{c|c|c|c|c|c|c}
		\hline
		Dataset&\multicolumn{3}{c|}{DDPM}&\multicolumn{3}{c}{Reverse sampling acceleration}\\
		\hline
		&$\rm{AUC}_{(P_f, P_d)}$&Steps&Time (in seconds)&$\rm{AUC}_{(P_f, P_d)}$&Steps&Time (in seconds)\\
		\hline
		MULL Gulfport&\bf{1.0000}&500&1.7158&0.9999&\bf{5}&\bf{0.3357}\\
		\hline
		HYDICE&\bf{0.9998}&500&4.1429&\bf{0.9998}&\bf{5}&\bf{0.3552}\\
		\hline
		Airport-4&\bf{0.9987}&500&4.7545&0.9986&\bf{5}&\bf{0.3845}\\
		\hline
		Urban-2&\bf{0.9969}&500&4.8850&\bf{0.9969}&\bf{5}&\bf{0.3826}\\
		\hline
	\end{tabular}
	\label{ablation_ddim}
\end{table*}
\subsubsection{Effectiveness of Reverse Sampling Acceleration} The adopted DDIM-based reverse sampling acceleration significantly boosts detection speed. To evaluate its effectiveness, we compare it with DDPM, as shown in Table \ref{ablation_ddim}. The results reveal that while detection performance remains stable or experiences a slight decline, the number of sampling steps is drastically reduced from 500 to 5. This reduction leads to a substantial decrease in time consumption across all four datasets. These findings demonstrate that the reverse sampling acceleration accelerates detection while maintaining performance, making the proposed SCDM-HTD more efficient and practical for real-world applications.
\section{Conclusion}
\label{conclusion}
This paper introduces a spectral-wise conditional diffusion model to address the issue of neglecting distribution learning in HTD. Unlike existing methods, our model reformulates HTD as a task of generating a detection map that learns the distributional discrepancy for effective background-target separation. Specifically, we first construct background and target confidence scores using linear mixing and spectral similarity computation. These confidence scores are treated as training samples, with the corresponding pixel spectra acting as guiding conditions, and are fed into the conditional diffusion model. The confidence scores are corrupted by adding Gaussian noise, and we propose a lightweight MLP-based CDNet to recover them. Furthermore, we introduce a composite loss function to enhance background-target separation. This loss integrates CEM loss to highlight the target while suppressing the background, and BCE loss to classify them into two distinct classes, thus improving detection robustness. Finally, we adopt a deterministic non-Markov chain to accelerate the reverse sampling process, requiring only a few steps to generate the detection map. Extensive experiments demonstrate that our model achieves outstanding performance in terms of objective metrics, subjective visual quality, and real-time requirements.

In future work, we plan to explore a generalized diffusion model pre-trained on public datasets, which can then be fine-tuned on specific datasets to improve both generalization and detection efficiency.
\bibliographystyle{IEEEtran}
\bibliography{SCDM-HTD}
\end{document}